\pdfoutput=1

\documentclass[11pt]{article}

\usepackage[]{ACL}

\usepackage{times}
\usepackage{latexsym}

\usepackage[T1]{fontenc}

\usepackage[utf8]{inputenc}

\usepackage{microtype}

\usepackage{inconsolata}
\usepackage{amsmath}
\usepackage{graphicx}
\usepackage{comment}
\usepackage{amssymb}
\usepackage[inkscapeformat=png]{svg}
\usepackage{paralist}
\usepackage{svg}
\usepackage{enumitem}
\title{RIFF: Learning to Rephrase Inputs \\
for Few-shot Fine-tuning of Language Models}

\author{Saeed Najafi \and
  Alona Fyshe \\
  Department of Computing Science, University of Alberta, Canada\\
  \texttt{\{snajafi,alona\}@ualberta.ca} \\}

\begin{document}
\maketitle
\begin{abstract}
Pre-trained Language Models (PLMs) can be accurately fine-tuned for downstream text processing tasks. Recently, researchers have introduced several parameter-efficient fine-tuning methods that optimize input prompts or adjust a small number of model parameters (e.g LoRA). 
In this study, we explore the impact of altering the \emph{input text} of the original task in conjunction with parameter-efficient fine-tuning methods. To most effectively rewrite the input text, we train a few-shot paraphrase model with a Marginal Maximum Likelihood objective. Using six few-shot text classification datasets, we show that enriching data with paraphrases at train and test time enhances the performance beyond what can be achieved with parameter-efficient fine-tuning alone. The code used for our experiments can be found at \url{https://github.com/SaeedNajafi/RIFF}.

\end{abstract}

\section{Introduction}
Multiple Pre-trained Language Models (PLMs), such as BERT~\cite{devlin-etal-2019-bert}, RoBERTa~\cite{DBLP:journals/corr/abs-1907-11692}, T5~\cite{DBLP:journals/corr/abs-1910-10683}, and GPT2~\cite{radford2019language}, have demonstrated remarkable performance when fine-tuned for downstream text processing tasks. PLM variants with less than 1 billion parameters are easier to train end-to-end with commodity hardware. However, very recent PLMs have been trained with a few hundred billion parameters, including PaLM-2 (540B)\cite{anil2023palm}, GPT3 (175B)\cite{brown2020language}, OPT (175B)\cite{zhang2022opt}, or Llama-2 (70B)\cite{touvron2023llama}. Training all parameters of these models end-to-end is not straightforward unless done with a dedicated cluster with specialized hardware.

In response, NLP research have developed effective techniques to control or alter the behavior of PLMs by updating the input context through prompt optimization~\cite{DBLP:journals/corr/abs-2107-13586} or adapting a few additional parameters within the network itself~\cite{DBLP:journals/corr/abs-2106-09685}.
However, current PLM control techniques have not considered altering the \textit{original input text} to improve the performance of the model.  Here, we investigate this idea by training a secondary smaller PLM to paraphrase the original input at train and test time, thus augmenting the existing data and improving model performance.

Our inspiration comes from interactions with young children.
Determining what a child knows is challenging because they can be sensitive to the wording of the question~\cite{childBook}. Adults are also influenced  by different wordings of a question.  For example, opinion polling has been found to be sensitive to the wording of questions~\cite{Broughton1995}. Just like we rephrase questions for humans, we should consider rephrasing input text while querying a PLM. For instance, while classifying the topic of a sentence, phrases related to time may be irrelevant and could be removed to simplify the modeling problem. Slight changes to wording could result in the model producing a correct prediction.

We explore the integration of paraphrased input texts during both the training and testing phases. At \emph{training time}, augmenting data through paraphrase generation has been shown to enhance performance while updating all parameters of the model~\cite{wei-zou-2019-eda, feng-etal-2021-survey, DBLP:journals/corr/abs-2106-07499, abaskohi-etal-2023-lm}. We broaden the scope of previous investigations by using paraphrase augmentation in tandem with recent prompt optimization and efficient tuning methods. At \emph{test time}, recent works have used ensemble predictions with various optimized prompts and tuned weights~\cite{izmailov2019averaging, li-etal-2023-making}. We further contribute to this line of work by incorporating ensemble predictions based on input paraphrases, again in concert with prompt optimization and efficient tuning techniques.

We start by pre-training a smaller language model on paraphrases generated by a large language model (i.e. ChatGPT or GPT3.5-turbo). Subsequently, we explore various training objectives for fine-tuning this paraphrase generator with feedback from the main task's language model. Our analysis shows that our proposed objective reduces hallucination in the generated paraphrases. Then, we experiment on six text classification datasets demonstrating that incorporating paraphrase augmentation during both training and testing phases enhances the performance of discrete/soft prompt optimization and efficient tuning techniques. In summary, our contributions are as follows:
\begin{itemize}[noitemsep,nolistsep]
\item We propose an efficient idea for Rephrasing Inputs for parameter-efficient Few-shot Fine-tuning of language models (RIFF) tested with recent prompt optimization and efficient tuning methods.
\item We conduct a comprehensive study on various learning objectives for fine-tuning a paraphrase generator using feedback from the main language model.
\item Our augmentation experiments on six text classification datasets reveal that paraphrase generation, when combined with prompt optimization and adaptation techniques is a simple yet effective approach to boost performance.
\end{itemize}

\section{Problem Formulation}
We focus on classification problems in Natural Language Understanding (NLU) tasks where we have access to a mini-batch of supervised training examples $B_{\text{supp}} = \{(x_i, y_i)\}_{i=1}^{N}$. Our goal is to update the parameter set $\theta_{\text{lm}}$ for a language model by maximizing the probability of the class label $y_i$ given the input $x_i$: $P_{\theta_{\text{lm}}} (y_i | x_i)$. Here, we augment $B_{\text{supp}}$ with semi-supervised examples. In particular, we generate $M$ paraphrases for each $x_i$ using the paraphrase generator $P_{\theta_{\text{par}}} (z_{i,j} | x_i)$, where $z_{i,j}$ represents the $j$-th paraphrase for the input $x_i$.  In the optimal case, this paraphrase will preserve semantic meaning but vary syntactic/lexical form. We then incorporate the generated paraphrases to create a new mini-batch of examples $B_{\text{s+p}} = B_{\text{supp}} \cup B_{\text{para}}$.  Using this augmented mini-batch, we optimize the following objective $J_{\theta_{\text{lm}}}$:
\begin{equation}
\sum_{i=1}^N \Bigl\{\log P_{\theta_{\text{lm}}} (y_i | x_i) + \\
\frac{1}{M} \sum_{j=1}^{M} \log P_{\theta_{\text{lm}}} (y_i | z_{i,j})\Bigr\}
\label{lmfp-augmentation-objective}
\end{equation}

To train the language model using Equation \ref{lmfp-augmentation-objective}, we need to update the parameter set $\theta_{\text{lm}}$. One approach would involve updating every parameter for the language model to optimize the training objective (referred to here as the "All-Finetuning" or \textit{AllTune} approach).  However, this method can be computationally intensive. As a result, we will explore the impact of paraphrase augmentation along with six other efficient baseline tuning techniques~\cite{pmlr-v97-houlsby19a} and prompt optimization~\cite{liu2021pretrain}.

We assume that each input $x$ or its paraphrase $z$ is preceded by the task instruction $p$, which is often specified in previous works. The task instruction, which we represent using the symbol $p$ to be consist with prompt optimization literature, serves as a parameter-free, gradient-free technique for enhancing the performance of the PLM across various downstream tasks~\cite{DBLP:journals/corr/abs-2005-14165, petroni-etal-2019-language, deng-etal-2022-rlprompt}. When using only the task instructions, no parameters for the language model are updated ($\theta_{\text{lm}}=\emptyset$), and zero-shot predictions are made solely on the evaluation data. We further investigate multiple language model tuning techniques while incorporating these task instructions into the input or its paraphrases.

\subsection{LM-Friendly Paraphrase Search}
\label{paraphrase-objectives}
Given a training example $(x, y)$, our objective is to assign the gold label $y$ to the input $x$  by maximizing the log likelihood $\log P(y|x)$. We leverage the fact that when $x$ is misclassified, there may exist paraphrases of the input $x$ that lead to the correct class prediction. These paraphrases should retain the semantic meaning of $x$ while exhibiting syntactic differences, akin to the way we rephrase things when we have been misunderstood. Thus, we generate paraphrases $z_{j}$ based on the input $x$, that enable the downstream language model to predict the correct label $y$ with greater confidence. Consequently, our data log likelihood is factorized into the following marginalization over the space of paraphrases, where $\theta_{\text{par}}$ and $\theta_{\text{lm}}$ represent the parameters for the paraphrase generator and the downstream language model, respectively:
\begin{multline}
J_{\theta_{\text{par}}} := \log P(y | x) = \log E_{z} [P(y, z| x)] \\ = \log \sum_{z} P_{\theta_{\text{par}}}(z | x) \times P_{\theta_{\text{lm}}}(y | z)
\label{lmfp-main-objective}
\end{multline}

To train the paraphrase generator and optimize the objective stated in Equation~\ref{lmfp-main-objective}, we explore four distinct learning aspects:
\begin{inparaenum}[(a)]
\item two methods for gradient approximation,
\item a reward normalization technique,
\item three decoding techniques for sampling paraphrases, and
\item two approaches to ensure grammatical integrity during paraphrase generation.
\end{inparaenum}
By combining these elements, we examine various learning approaches to refine the paraphrase generator with the aid of the downstream language model. In the subsequent paragraphs, we will describe our suggested options for each aspect.

\textbf{Gradient Approximation}:
\noindent
Text generation can be reformulated as an episodic reinforcement learning problem where an agent (i.e. a paraphrase generator) generates tokens (i.e. takes actions) one step at a time until reaching the end of the episode (i.e. selecting the end of sequence token). Therefore, for a given training example $(x, y)$ and its paraphrase $z$, we define the terminal reward (i.e. goodness) for $z$ as $R(z) = \log P_{\theta_{\text{lm}}} (y | z)$. When approximating the gradient vector of objective~\ref{lmfp-main-objective} concerning $\theta_{\text{par}}$, we propose two strategies. These include: (i) \emph{Marginal Maximum Likelihood (MML)} and (ii) approximating the gradient vector of the paraphrase model via the \emph{Policy Gradient (PG)} theorem. Notably, gradient updates using these two methods exhibit a close relationship, with the main difference lying in the posterior coefficient utilized to score each sample~\cite{guu-etal-2017-language}. We can recast the main objective presented in equation \ref{lmfp-main-objective} into the following equation representing the expected reward:
\begin{equation}
J_{\theta_{\text{par}}} := \log E_{z \, \sim \, P_{\theta_{\text{par}}}(.|x)} [e^{R(z)}]
\label{lmfp-expect-objective}
\end{equation}

Given each input $x$, if we extract paraphrase samples from $P_{\theta_{\text{par}}}(.|x)$ and approximate the expectation in $J_{\theta_{\text{par}}}$ via numerical summation, we optimize the objective using MML estimation. This process results in the following gradient update:
\begin{multline}
\nabla J^{\text{MML}}_{\theta_{\text{par}}} := \nabla_{\theta_{\text{par}}} \log E_{z} [e^{R(z)}] \\ =
\sum^{M}_{j=1} \phi^{\text{MML}}(z_{j}) \times \nabla_{\theta_{\text{par}}} \log P_{\theta_{\text{par}}}(z_{j}|x) \\
\phi^{\text{MML}}(z_{j}) = \frac{P_{\theta_{\text{par}}}(z_{j}|x) \times e^{R(z_{j})}}{\sum^{M}_{j^{'}=1} P_{\theta_{\text{par}}}(z_{j^{'}}|x) \times e^{R(z_{j^{'}})}}
\label{mml-objective}
\end{multline}

By introducing the $\log$ inside the expectation (applying Jensen's inequality), we can optimize a surrogate lower bound for the objective presented in equation ~\ref{lmfp-expect-objective}, resulting in the following policy gradient approximation~\cite{10.5555/3009657.3009806}:
\begin{multline}
\nabla J^{\text{PG}}_{\theta_{\text{par}}} := \nabla_{\theta_{\text{par}}} E_{z} [R(z)] \\ =
\sum^{M}_{j=1} \phi^{\text{PG}}(z_{j}) \times \nabla_{\theta_{\text{par}}} \log P_{\theta_{\text{par}}}(z_{j}|x) \\
\phi^{\text{PG}}(z_{j}) = P_{\theta_{\text{par}}}(z_{j}|x) \times R(z_{j})
\label{pg-objective}
\end{multline}

\textbf{Reward Normalization}:
\noindent
For our secondary learning aspect, we can either utilize the basic reward, denoted as $R(z_{j})$, or normalize the rewards among the paraphrases of a given input $x$. This process of normalization is particularly useful because it prevents the training of the paraphrase generator with rewards of varying magnitudes, as different training examples are not equally challenging for the language model. Prior research suggests that such normalization of rewards can significantly enhance the performance of text generators across a variety of tasks \cite{guo-etal-2022-efficient}. The normalized reward $R^{n}$ is defined as follows:
\begin{multline}
R^{n}(z_{j}) = \frac{R(z_{j}) - \mu_{j}}{\sigma_{j}}, \mu_{j} = \frac{1}{M} \sum^{M}_{j=1} R(z_{j}) \\
\sigma^{2}_{j} = \frac{1}{M} \sum^{M}_{j=1} (R(z_{j}) - \mu_{j})^2
\label{normal-reward}
\end{multline}

\textbf{Decoding Techniques}:
\noindent
To train the paraphrase generator, we use both the MML and PG gradient estimations which necessitates drawing $M$ samples from the paraphrase generator. We implement three decoding techniques for this purpose. Firstly, we utilize diverse beam search decoding~\cite{Vijayakumar_Cogswell_Selvaraju_Sun_Lee_Crandall_Batra_2018} to gather these $M$ paraphrases. Secondly, in order to thoroughly explore the paraphrase space, we alternatively collect the $M$ paraphrases using nucleus (top-p) sampling~\cite{holtzman2020curious}. For the top-p sampling, we establish a sampling threshold of $p=0.99$, at which we collect the minimal set of tokens from the vocabulary with a cumulative probability of at least $0.99$. We then re-sample tokens from this set. And thirdly, during the training phase we blend diverse beam search and top-p sampling. Here, we initially sample $M$ paraphrases using both methods, then combine the top $M/2$ samples from each output to construct our final $M$ samples. During the test phase, we only use diverse beam search.

\textbf{Grammatical Integrity}:
\noindent
We describe three distinct modeling techniques for both the MML and PG gradient estimations: On-policy learning, Off-policy learning and KL-penalized On-policy (KLOn) learning.

As we are sampling paraphrases from $P_{\theta_{\text{par}}}(z_{j}|x)$ and updating $\theta_{\text{par}}$ using these samples, the paraphrase generator may start generating ungrammatical text during this default on-policy learning setting. Similar instances of degenerate generation have been reported in tasks like question generation~\cite{najafi-fyshe-2023-weakly} and program synthesis~\cite{NEURIPS2018_f4e369c0}.

To mitigate degenerate paraphrase generation, we experiment with off-policy sampling. Here, we maintain a fixed sampling module $P_{\text{fixed}}(z_{j}|x)$ for sample selection, then update the main paraphrase generator $P_{\theta_{\text{par}}}(z_{j}|x)$ within the frameworks of objectives \ref{mml-objective} and \ref{pg-objective}. Consequently, with these off-policy samples, the posterior coefficients incorporate the importance sampling ratio $s(z_{j}) = \frac{P_{\theta_{\text{par}}}(z_{j}|x)}{P_{\text{fixed}}(z_{j}|x)}$
\begin{multline}
\phi^{\text{PG}}_{\text{off}}(z_{j}) = s(z_{j}) \times R(z_{j})\\
\phi^{\text{MML}}_{\text{off}}(z_{j}) = \frac{s(z_{j}) \times e^{R(z_{j})}}{\sum^{M}_{j^{'}=1} s(z_{j^{'}}) \times e^{R(z_{j^{'}})}}
\label{off-pg-mml-objective}
\end{multline}

Our next solution for degenerate paraphrases involves imposing a penalty in the training objective if the samples drawn from the current paraphrase generator, $P_{\theta_{\text{par}}}(z|x)$, deviate from those of the pre-trained paraphrase generator. We can implement this penalty as a KL-divergence penalty between the distributions of paraphrases produced by the current model and the pre-trained one. To integrate this penalty we define the following new objective for $\theta_{\text{par}}$:
\begin{equation}
J^{\text{KLOn}}_{\theta_{\text{par}}}
:= \log E_{z} [e^{R(z)}] - \beta E_{z} [\log s(z)]
\label{lmfp-expect-KLOn-objective}
\end{equation}

Building upon the previously approximated MML and PG gradients, we can now derive the following regularized gradient vector with respect to $\theta_{\text{par}}$. Please note that $\beta$ is a hyper-parameter in this context:
\begin{multline}
\nabla J_{\theta_{\text{par}}} - \beta E_{z} [(\log s(z) + 1) \nabla \log P_{\theta_{\text{par}}} (z | x)],\\
z \sim P_{\theta_{\text{par}}}(.|x)
\label{lmfp-expect-KLOn-gradient}
\end{multline}

Note that the KL penalty can be interpreted as the sum of a grammar reward, denoted by $\log P_{\text{fixed}}(z|x)$, and an entropy regularization term over $P_{\theta_{\text{par}}} (z | x)$. The entropy regularization aids in the diverse exploration of the search space~\cite{DBLP:journals/corr/MnihBMGLHSK16}, while the grammar reward discourages the learning of ungrammatical samples.

\subsection{Ensemble Inference}
\label{ensemble-inference}
After optimizing Equation \ref{lmfp-main-objective} and fine-tuning our paraphrase generator, we generate weakly-supervised examples for inclusion in Equation \ref{lmfp-augmentation-objective} to train our downstream language model.

To predict the class label of a test example, we could either use our fine-tuned language model to predict the class label based on the original input $x$, or adopt an ensemble approach. For the latter, for a given $x$, we generate $M$ paraphrases using our fine-tuned paraphrase generator. We then average the prediction scores for a potential class label across the $M+1$ values according to Equation~\ref{lmfp-augmentation-objective} to predict the class label for that input example $x$. This aligns with our earlier assumption that some paraphrases could be easier for the language model to predict the correct class label. During data augmentation for the language model, we select the validation set's best model according to this ensemble prediction.

\section{Experiments}

\subsection{Setup}

\textbf{Pre-trained Models}:
\noindent

For paraphrase generation, we employ a T5-base model~\cite{DBLP:journals/corr/abs-1910-10683} which has been trained on paraphrases generated by ChatGPT (i.e. version GPT3.5-turbo). These output paraphrases were generated for input texts from various datasets, including Quora paraphrase questions, texts from SQUAD 2.0, and the CNN news dataset~\cite{chatgpt_paraphrases_dataset}. To create this training data, ChatGPT generated five paraphrases for each input, which were then used as the target for the T5-base model. The weights for this model are publicly available\footnote{\tiny \url{https://huggingface.co/humarin/chatgpt_paraphraser_on_T5_base}}. In our experiments, this model was able to generate more diverse paraphrases compared to other public pre-trained models.

For our main language model, we use the RoBERTa-large model pre-trained with the Masked Language Modeling (MLM) objective~\cite{DBLP:journals/corr/abs-1907-11692}, which has demonstrated strong performance on NLU tasks. Our proposed learning framework can be readily extended to other paraphrase generators or backbone language models.

\noindent
\textbf{Datasets}: Inspired by prior work~\cite{gao-etal-2021-making, deng-etal-2022-rlprompt}, we experiment on six classification tasks in the few-shot setting. These include sentiment classification tasks such as the binary sentiment datasets SST2~\cite{socher-etal-2013-recursive}, CR~\cite{10.1145/1014052.1014073}, and MR~\cite{pang-lee-2005-seeing}. We also experiment on the 5-label sentiment dataset SST5~\cite{socher-etal-2013-recursive}, the question type classification dataset TREC~\cite{10.1145/345508.345577}, and the topic classification dataset AGNews~\cite{DBLP:journals/corr/ZhangZL15}. The number of classes per dataset, as well as the used instructions are outlined in Appendix~\ref{task-instruct-input-format:appendix}. Instructions and class verbalizers are based on previous work~\cite{deng-etal-2022-rlprompt} in prompt optimization. Detailed information about the specific learning rates for each LM technique along with other hyper-parameters can be found in Appendix~\ref{training-details-extra:appendix}.

\subsection{Few-shot Paraphrase Fine-Tuing}
\label{final-RIFF-result}

\begin{table*}
\centering
\caption{The average accuracy of the best performing validation checkpoint in the 128-shot SST2 classification task for both the on-policy and KLOn learning techniques. Highest performance per column bolded. Last column reports the macro-average among each row. Numbers in parentheses ($\sigma$ $\mid$ $m$): $\sigma$ represents the standard deviations for the reported means across five train/validation splits; $m$ reports the average accuracy of all the validation checkpoints as we monitor robustness during the training trajectory. The suffix `-$Z$' denotes models trained with reward normalization.}
\begin{tabular}{ c | c c c | c c c | c}
\hline
\small Learning & \multicolumn{3}{c|}{On-Policy} & \multicolumn{3}{c|}{KLOn} & AVG \\
\small Technique & \small{Top-P} & \small{Beam} & \small{Mixed} & \small{Top-P} & \small{Beam} & \small{Mixed} & \\
\hline
\small No Tuning & \small67.5 & \small67.5 & \small67.5 & \small67.5 & \small67.5 & \small67.5 & \small67.5\\
\hline
\small PG & \small67.9 \tiny (2.3 $\mid$ 53.2) & \small68.0 \tiny (1.4 $\mid$ 52.0) & \small67.9 \tiny (2.0 $\mid$ 52.4) & \small68.5 \tiny (1.2 $\mid$ 67.4) & \small68.3 \tiny (1.9 $\mid$ 67.4) & \small69.1 \tiny (1.4 $\mid$ 68.2) & \small68.3 \tiny (1.7 $\mid$ 60.1)\\
\small PG-Z & \small\textbf{71.3} \tiny (2.1 $\mid$ 63.8) & \small\textbf{70.2} \tiny (1.7 $\mid$ 68.6) & \small\textbf{71.2} \tiny (1.3 $\mid$ 66.6) & \small68.9 \tiny (1.3 $\mid$ 67.6) & \small68.8 \tiny (1.4 $\mid$ 67.9) & \small69.8 \tiny (1.3 $\mid$ 68.5) & \small70.0 \tiny (1.5 $\mid$ 67.2) \\
\hline
\small MML & \small69.6 \tiny (2.1 $\mid$ 68.5) & \small69.1 \tiny (1.8 $\mid$ 67.6) & \small69.8 \tiny (2.0 $\mid$ 68.6)& \small\textbf{69.5} \tiny (2.4 $\mid$ 68.2)& \small69.9 \tiny (1.9 $\mid$ 69.0)& \small70.5 \tiny (3.0 $\mid$ 68.9) & \small69.7 \tiny (2.2 $\mid$ 68.5) \\
\small MML-Z & \small70.3 \tiny (2.7 $\mid$ 69.1) & \small\textbf{70.2} \tiny (2.0 $\mid$ 68.9) & \small70.2 \tiny (2.3 $\mid$ 69.1) & \small68.9 \tiny (1.8 $\mid$ 67.9) & \small\textbf{70.3} \tiny (1.7 $\mid$ 69.0) & \small\textbf{70.6} \tiny (2.5 $\mid$ 68.9) & \small\textbf{70.1} \tiny (2.2 $\mid$ 68.8) \\
\hline
\end{tabular}
\label{results-on-ppo}
\end{table*}

As discussed in Section~\ref{paraphrase-objectives}, there are four learning aspects to be considered when fine-tuning our paraphrase generator for the downstream language model. We conduct an extensive set of experiments in the 128-shot setting for the SST2 binary sentiment classification task.

\begin{figure}[h]
\begin{center}
\includegraphics[width=0.4\textwidth]{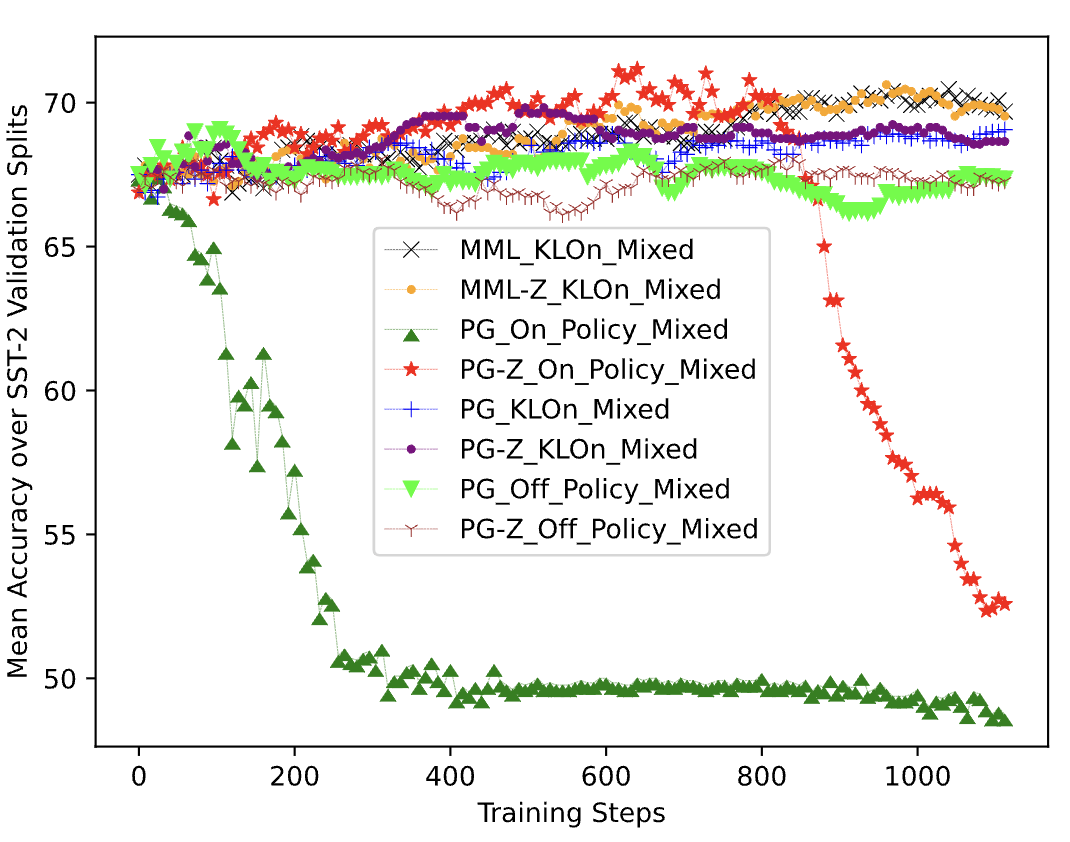}
\end{center}
\caption{Average ensemble accuracy over five validation splits in the 128-shot SST2 classification task. PG gradient estimation is not robust during the training trajectory while doing on-policy learning.}
\label{pg_divergence}
\end{figure}

We randomly select 128 training examples for each unique label within the dataset. An equal number of examples are gathered to form an internal validation set. We create five train/validation splits using the arbitrarily chosen random seeds. We train the models for 1120 training steps with the batch size of 8 (i.e. 35 epochs). As we are training the models, we evaluate the performance of 140 weight checkpoints per model on the validation splits (i.e one checkpoint per 8 training steps). We examine the mean accuracy, which is averaged over the five validation splits. Despite the ensembling approach described in Section~\ref{ensemble-inference}, to accurately capture the quality of the generated paraphrases, we exclude the original input $x$ when computing the ensemble accuracy on the validation splits.

We assess the impact of reward normalization in the context of on-policy, off-policy, and KL-penalized on-policy (KLOn) learning, considering both PG and MML gradient estimations. Table~\ref{results-on-ppo} lists the best performance out of all the checkpoints evaluated on the validation splits, which is further averaged over five validation splits. With both PG and MML gradient estimations, reward normalization is boosting the performance across the three text decoding techniques for both on-policy and KLOn learning techniques (see `AVG' column in Table~\ref{results-on-ppo}). Conversely, reward normalization is not improving performance with off-policy learning (follow discussion in Appendix~\ref{training-paraphrase-extra:appendix} and see Table~\ref{maximum-curves-off})

Table~\ref{results-on-ppo} verifies that MML gradient estimation outperforms PG gradient estimation on average across three decoding techniques for both on-policy and KLOn learning techniques. The highest accuracy is achieved by `PG-Z' with on-policy learning and top-p decoding, however it is not robust during the entire training trajectory. Figure~\ref{pg_divergence} shows that PG gradient estimation is not robust throughout the training trajectory, which causes the paraphrase generator to produce nonsensical paraphrases.  This results in downstream classification performance on par with random guessing. In contrast, off-policy and KLOn learning circumvent this divergence. MML gradient estimation maintains robustness throughout the training phase. In Table~\ref{results-on-ppo}, we also report the average accuracy of all the checkpoints as we are training the models (numbers in parentheses). The learning technique `MML-Z' is more robust during the training trajectory compared to `PG-Z'.

Upon investigating various elements of our learning objectives for fine-tuning the paraphrase generator, the combination that delivers the best performance across the validation splits, which is also robust during the entire training trajectory, includes: \textbf{MML} gradient approximation, \textbf{KLOn} learning, \textbf{mixed decoding} for sample generation, and finally applying \textbf{reward normalization}. We name this combined approach our proposed RIFF algorithm.

\subsection{Paraphrase Quality Analysis}
\label{Paraphrase Quality Analysis}
We investigate the impact of the RIFF algorithm on the quality of the paraphrases in the 128-shot setting across three classification tasks SST2, SST5 and AGNews. To evaluate the quality of the paraphrases, we report the following five metrics:

\noindent
\textbf{Grammar ($GR$)}:
We evaluate grammar by calculating the perplexity score averaged across the dataset using the GPT-2-Large model~\cite{Radford2019LanguageMA}. A low $GR$ score indicates more grammatical texts.

\noindent
\textbf{Lexical Diversity ($LD$)}:
To assess how paraphrases differ lexically from the \emph{original} input text, we calculate the unigram and bigram Rouge scores between each paraphrase and the original input text~\cite{lin-2004-rouge}. We then report $1 - \frac{(Rouge\_1 + Rouge\_2)}{2}$ as our lexical diversity metric. A higher $LD$ score indicates greater lexical difference compared to the original text.

\noindent
\textbf{Pair-wise Lexical Diversity ($PLD$)}:
To assess the lexical diversity among the set of \emph{paraphrases} for a given original input text, we calculate $LD$ scores for every pair of paraphrases for an input text, and report the average. A higher $PLD$ score indicates greater diversity among the paraphrases for a specific input text.

\noindent
\textbf{Semantic Similarity ($SS$)}:
To assess how semantically similar paraphrases are to the original input text, we compute the BERTScore's F1 metric~\cite{Zhang*2020BERTScore:} between each paraphrase and the original input text using the BERT-Large model. A higher $SS$ score signifies more semantic similarity with the original text.

\noindent
\textbf{Factual Consistency ($FC$)}:
To measure hallucination in the generated paraphrases with respect to the original input text, we rely on a publicly available factual consistency metric. The model has been trained for textual entailment and summarization datasets with samples annotated for factual consistency\footnote{{\tiny \url{https://huggingface.co/vectara/hallucination_evaluation_model}}}.

\begin{table}
\centering
\caption{Average metrics on the test sets to evaluate the quality of paraphrases across three classification tasks: SST2, SST5, and AGNews datasets. The metrics are further averaged across five training folds for the \textit{FinPara} method. \textit{OrigIn}: Represents the original task inputs. \textit{PrePara}: Corresponds to task inputs obtained from the pre-trained paraphraser. \textit{FinPara}: Indicates task inputs from the finetuned paraphraser in the $128$-shot setting. We scale scores into the range of [0-100], except for $GR$. Better performance per column bolded.}
\begin{tabular}{ c | c c c c c}
\hline
Input Type & $GR$ & $LD$ & $PLD$ & $SS$ & $FC$ \\
\hline
\small \textit{OrigIn} & \small 198 & n/a & n/a & n/a & n/a \\
\small \textit{PrePara} &  \small \textbf{143} & \small \textbf{60.5} & \small \textbf{61.6} & \small 70.4 & \small 77.3 \\
\small \textit{FinPara} & \small 162 & \small 50.6 & \small 53.3 & \small \textbf{74.7} & \small \textbf{79.1} \\
\hline
\end{tabular}
\label{paraphrase-quality}
\end{table}

We present the metrics for the datasets in Table~\ref{paraphrase-quality}. Compared to a model that was not fine-tuned for this task (PrePara), our RIFF algorithm has reduced the diversity among the generated paraphrases and their lexical variation compared to the original input text. This outcome aligns with our search-learn objective that prioritizes high-scoring paraphrases over others. RIFF has contributed to higher semantic similarity compared to the original input. Interestingly, the perplexity of the generated paraphrases after fine-tuning with our objective is still low, demonstrating the grammatical accuracy of these paraphrases. The example paraphrases shown in Appendix~\ref{example-paraphrases:appendix} illustrate that RIFF reduces hallucination in the generated paraphrases, which may contribute to the lower $LD$ score but higher $SS$ score with respect to the original input text. A higher $FC$ score as shown in Table~\ref{paraphrase-quality} verifies that the RIFF objective has reduced hallucination in the generated paraphrases.

\subsection{Paraphrases for Few-shot LM Tuning}

\begin{table*}
\centering
\caption{Average accuracy on the standard evaluation sets for the 16-shot text classification. Numbers in parentheses are standard deviations across the five train/validation folds. The last column is the micro averaged performance across the datasets. Highest performance per dataset bolded, second highest underlined. $\dagger$: the average 16-shot \textit{AllTune} results with automatically searched templates~\cite{gao-etal-2021-making}. $\star$: reported results for RoBERTa-large using in-context learning~\cite{gao-etal-2021-making}.}
\begin{tabular}{c | c | c | c | c | c | c || c }
\hline
Tuning Method & SST2 & SST5 & CR & MR & TREC & AGN & AVG \\
\hline
\small No Tuning & \small84.6 & \small31.0 & \small77.8 & \small81.3 & \small27.6 & \small 51.5 & \small58.6 \\
\small ICL$\star$ & \small84.8 & \small30.6 & \small87.4 & \small80.5 & \small26.2 & \small - & \small66.9 \\
\small RLPrompt & \small92.5 & \small41.4 & \small89.5 & \small\underline{87.1} & \small60.5 & \small 80.2 & \small77.7 \\
\small LM-BFF$\dagger$ & \small92.3 & \small49.2 & \small89.0 & \small85.5 & \small\underline{88.2} & \small - & \small78.5 \\
\hline
\small \textit{\textit{ClsTune}} & \small72.6 \tiny (2.4) & \small34.4 \tiny (2.6) & \small71.4 \tiny (2.7) & \small67.3 \tiny (2.8) & \small74.8 \tiny (3.4) & \small 81.7 \tiny (1.6) & \small70.9  \tiny (2.2) \\
\tiny+RIFF (train)  & \small72.5 \tiny (3.4) & \small33.9 \tiny (3.7) & \small68.3 \tiny (3.9) & \small70.3 \tiny (0.9) & \small75.8 \tiny (1.7) & \small 84.0 \tiny (0.9) & \small71.9  \tiny (2.0)\\
\tiny+RIFF (train+test)  & \small74.0 \tiny (3.3) & \small35.0 \tiny (3.6) & \small71.1 \tiny (4.4) & \small72.0 \tiny (1.7) & \small76.8 \tiny (2.9) & \small 84.9 \tiny (0.9) &\small73.3  \tiny (2.1)\\
\hline
\small \textit{\textit{\textit{\textit{GS}}}} & \small85.5 \tiny (1.3) & \small37.0 \tiny (4.2) & \small80.2 \tiny (2.3) & \small83.0 \tiny (1.8) & \small45.3 \tiny (13.1) & \small 82.0 \tiny (1.4) & \small75.0 \tiny (2.3) \\
\tiny+RIFF (train) & \small86.4 \tiny (1.9) & \small37.8 \tiny (3.3) & \small82.7 \tiny (1.3) & \small84.7 \tiny (2.1) & \small51.0 \tiny (8.6) & \small 81.0 \tiny (2.4) & \small75.4  \tiny (2.5)\\
\tiny+RIFF (train+test) & \small87.3 \tiny (2.0) & \small38.2 \tiny (3.5) & \small85.1 \tiny (1.9) & \small84.7 \tiny (1.9) & \small52.4 \tiny (7.7) & \small 83.3 \tiny (1.5) & \small77.0  \tiny (2.1)\\
\hline
\small \textit{\textit{SpTune}} & \small89.7 \tiny (3.7)& \small39.4 \tiny (6.2) & \small82.4 \tiny (2.8) & \small86.1 \tiny (2.2) & \small35.2 \tiny (2.7) & \small 82.0 \tiny (2.6) & \small76.1  \tiny (3.2)\\
\tiny+RIFF (train) & \small91.2 \tiny (2.2) & \small44.5 \tiny (4.2) & \small84.6 \tiny (1.9) & \small86.1 \tiny (0.7) & \small38.4 \tiny (4.3) & \small 84.0 \tiny (1.9) & \small78.3  \tiny (2.2)\\
\tiny+RIFF (train+test) & \small91.6 \tiny (2.3) & \small45.1 \tiny (4.1) & \small86.2 \tiny (1.8) & \small86.6 \tiny (0.8) & \small38.4 \tiny (4.0) & \small 86.0 \tiny (1.0) & \small79.7  \tiny (1.7)\\
\hline
\small \textit{\textit{HTune}} & \small87.4 \tiny (2.3) & \small37.4 \tiny (2.0) & \small84.0 \tiny (2.8) & \small83.1 \tiny (1.8) & \small62.4 \tiny (7.4) & \small 81.4 \tiny (1.4) & \small76.0  \tiny (2.0) \\
\tiny+RIFF (train) & \small88.1 \tiny (1.7) & \small40.3 \tiny (1.9) & \small84.5 \tiny (1.5) & \small83.4 \tiny (2.8) & \small70.7 \tiny (4.8) & \small 83.4 \tiny (0.9) &\small77.8  \tiny (1.6)\\
\tiny+RIFF (train+test) & \small89.1 \tiny (1.2) & \small40.4 \tiny (1.8) & \small86.4 \tiny (0.8) & \small83.1 \tiny (3.7) & \small71.6 \tiny (5.9) & \small 85.2 \tiny (1.1) &\small79.0  \tiny (1.6)\\
\hline
\small \textit{\textit{InTune}} & \small91.5 \tiny (1.2) & \small42.3 \tiny (4.2) & \small87.3 \tiny (2.0) & \small84.0 \tiny (2.5) & \small67.7 \tiny (5.8) & \small 83.8 \tiny (2.2) & \small78.9  \tiny (2.5)\\
\tiny+RIFF (train) & \small92.6 \tiny (0.3) & \small43.2 \tiny (1.9) & \small87.5 \tiny (1.8) & \small85.9 \tiny (2.3) & \small63.8 \tiny (5.3) & \small 85.6 \tiny (0.8) &\small80.2  \tiny (1.3) \\
\tiny+RIFF (train+test) & \small93.1 \tiny (0.6) & \small43.9 \tiny (2.3) & \small89.0 \tiny (1.8) & \small86.0 \tiny (2.4) & \small69.6 \tiny (6.3) & \small 86.9 \tiny (0.4) &\small81.3  \tiny (1.3)\\
\hline
\small \textit{\textit{AllTune}} & \small93.1 \tiny (0.4) & \small48.0 \tiny (1.0) & \small89.2 \tiny (0.8) & \small\textbf{87.3} \tiny (3.1) & \small87.2 \tiny (3.8) & \small \underline{87.7} \tiny (0.5) & \small\underline{83.0}  \tiny (1.0)\\
\tiny+RIFF (train) & \small\underline{93.6} \tiny (1.3) & \small\underline{50.6} \tiny (1.0) & \small\underline{90.2} \tiny (1.5) & \small85.8  \tiny (2.3) & \small84.2 \tiny (4.9) & \small 87.2 \tiny (0.6) & \small\underline{83.0}  \tiny (1.2)\\
\tiny+RIFF (train+test) & \small\textbf{93.8} \tiny (1.2) & \small\textbf{51.2} \tiny (1.6) & \small\textbf{91.0} \tiny (1.6) & \small85.5 \tiny (2.3) & \small84.4 \tiny (4.9) & \small 87.2 \tiny (0.6) & \small\textbf{83.2}  \tiny (1.3)\\
\hline
\small \textit{\textit{LoRA}} & \small92.5 \tiny (1.8) & \small48.1 \tiny (1.8) & \small88.6 \tiny (2.0) & \small86.0 \tiny (2.6) & \small\textbf{89.3} \tiny (2.2) & \small 87.3 \tiny (0.5) & \small82.6  \tiny (1.3)\\
\tiny+RIFF (train) & \small92.7 \tiny (1.8) & \small48.0 \tiny (2.3) & \small87.5 \tiny (1.5) & \small85.1 \tiny (2.9) & \small84.8 \tiny (2.7) & \small 87.6 \tiny (0.3) & \small82.3  \tiny (1.3)\\
\tiny+RIFF (train+test) & \small93.1 \tiny (1.2) & \small49.2 \tiny (2.0) & \small89.0 \tiny (1.1) & \small85.4 \tiny (2.8) & \small85.9 \tiny (3.6) & \small \textbf{87.9} \tiny (0.3) &\small82.9  \tiny (1.1)\\
\hline
\end{tabular}
\label{RIFF-vs-main-16-shot}
\end{table*}

Our primary hypothesis is that various LM tuning techniques could benefit from diverse views of the original input text. To test this hypothesis, we fine-tuned our paraphrase generators in a 16-shot classification setup using the RIFF algorithm. Subsequently, we fine-tuned the downstream classification model in the same 16-shot setting, while introducing $M=8$ paraphrases as per the objective outlined in Equation~\ref{lmfp-augmentation-objective}. For evaluation, we used the best model from the validation set to make predictions on standard evaluation splits, following the ensemble approach described in Section~\ref{ensemble-inference}.
For consistency with prior research, we used the random dataset splits provided by RLPrompt~\cite{deng-etal-2022-rlprompt}, aligning with the random seeds used by LM-BFF~\cite{gao-etal-2021-making}.

We study the effect of paraphrases on seven language model tuning techniques. The first technique \textit{\textit{AllTune}} updates every parameter in the network. Another technique, \textit{\textit{\textit{\textit{GS}}}}, is based on AutoPrompt~\cite{shin-etal-2020-autoprompt} for discrete prompt optimization. The technique \textit{\textit{SpTune}}~\cite{lester-etal-2021-power} learns soft prompt vectors, and \textit{\textit{LoRA}}~\cite{DBLP:journals/corr/abs-2106-09685} is a recent adaptation technique. Additionally, we investigate \textit{\textit{ClsTune}}, which trains a classifier on top of the language model, \textit{\textit{InTune}}, which updates all of the input embedding table, and \textit{\textit{HTune}}, which only updates the language modeling head in the Transformer architecture. A detailed description of these techniques is discussed in Appendix~\ref{lm-tuning-techniques:appendix}.

Table~\ref{RIFF-vs-main-16-shot} illustrates the average accuracy on standard test sets across six text classification datasets. The reported scores correspond to seven distinct LM tuning techniques: \textit{\textit{ClsTune}}, \textit{\textit{GS}}, \textit{\textit{SpTune}}, \textit{\textit{HTune}}, \textit{\textit{InTune}}, \textit{\textit{AllTune}}, and \textit{\textit{LoRA}}.

Recent prompt optimization techniques like \textit{\textit{\textit{GS}}} and \textit{SpTune} significantly benefit from paraphrase augmentation during training, with \textit{SpTune} demonstrating the most dramatic improvement (2.2\% average accuracy increase). While \textit{LoRA} already outperforms these techniques~\cite{DBLP:journals/corr/abs-2106-09685}, paraphrase augmentation further enhances its efficiency in learning adaptation matrices, leading to average accuracy gains of 0.2\% on SST2 and 0.3\% on AGNews. When coupled with ensemble predictions, denoted in rows with ``+RIFF (train+test)'', all LM tuning techniques see improvements from the generated paraphrases.

\subsection{Paraphrase Robustness Analysis}

\begin{figure}[h]
\begin{center}
\includegraphics[width=0.4\textwidth]{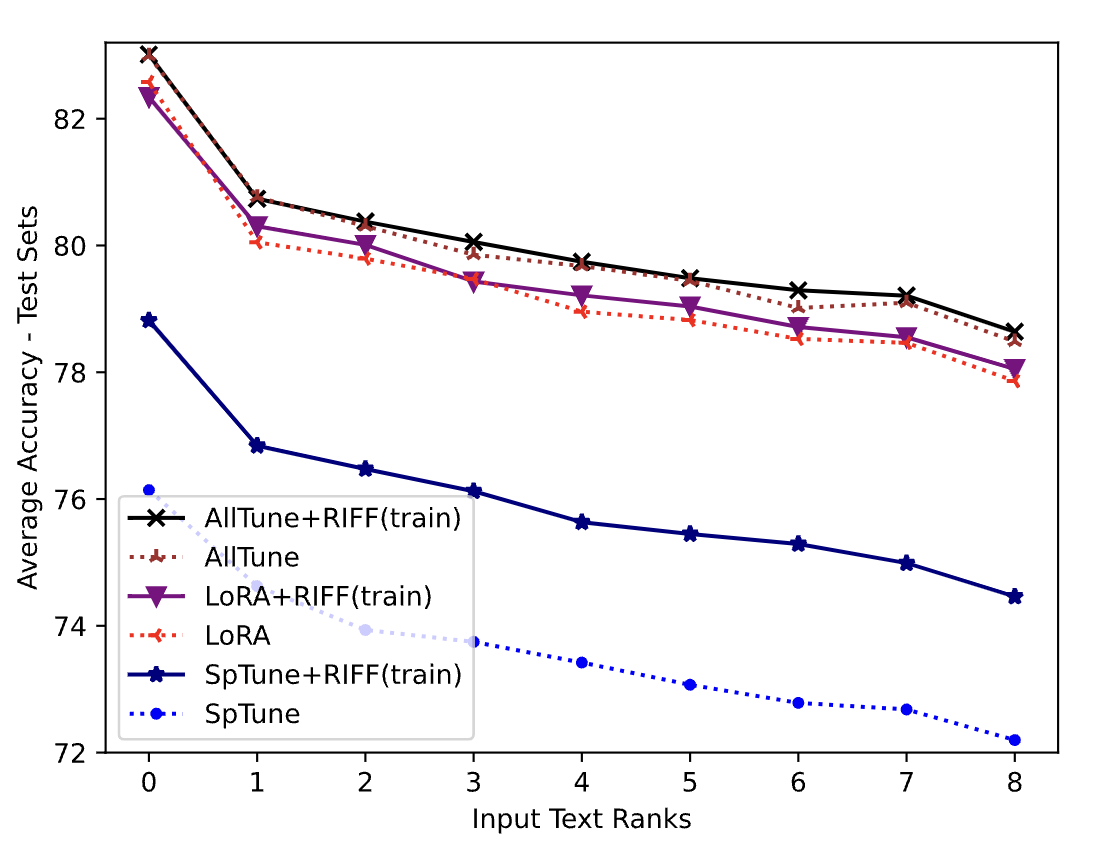}
\end{center}
\caption{Average test accuracy across six classification datasets. Input at rank 0 represents the original test input, while the remaining eight inputs are top-ranked paraphrases generated by our fine-tuned paraphrase model.}
\label{paraphrase-robustness}
\end{figure}

We generate $M = 8$ paraphrases for each test input text and investigate the average performance of the fine-tuned language models on each of these paraphrases compared to the original input text. Figure~\ref{paraphrase-robustness} illustrates the average test accuracy over six classification datasets for \textit{AllTune}, \textit{LoRA}, and \textit{SPTune}, which are popular LM tuning techniques in NLP. The original input texts are denoted with inputs at rank 0, whereas rank\_$i$ ($i \in \{1, 2, 3, 4, 5, 6, 7, 8\}$) are the paraphrases returned by the diverse beam search. We observe that the LM tuning techniques are not robust to paraphrases. Our paraphrase augmentation during training increases paraphrase robustness. \textit{SPTune} observes more significant improvement in robustness from paraphrase augmentation during training.

\subsection{Limitations Discussion}
Our paraphrase generator is pre-trained on semi-supervised paraphrases given by a truly large language model (i.e. ChatGPT). Although these large models are capable of generating high quality paraphrases for the English language. It is not clear if these semi-supervised paraphrases are available for other languages.

In terms of training overhead for our method, once the paraphrase model is fine-tuned, augmenting the mini-batches with paraphrases has minimal impact on training time in the downstream classification task. This is because we only need to perform inference with the paraphrase model once for the training examples to generate their paraphrases before the first epoch. The generated paraphrases are then cached for subsequent training epochs while fine-tuning the downstream language model.

In contrast, fine-tuning the paraphrase model using the RIFF objective requires generating new paraphrases each epoch, which prevents us from caching paraphrases across epochs. This fine-tuning step is about eight times slower than maximum-likelihood training, which benefits from having ground-truth paraphrases available.  Generating new samples and scoring them is a standard procedure in Reinforcement Learning.

\section{Related Works} 
To improve prompt optimization and efficient tuning techniques for LMs, we incorporate the generated paraphrases into the training mini-batches. Paraphrase generation represents just one technique of data augmentation. For a comprehensive overview of diverse data augmentation techniques for NLP tasks, we direct interested readers to a recent survey by \citet{DBLP:journals/corr/abs-2106-07499}.

A recent work for few-shot prompt-based learning helps contrastive training by paraphrasing the inputs~\cite{abaskohi-etal-2023-lm}. Our work proposes an objective to further fine-tune the paraphrase generator distilled from an LLM that reduces hallucination. Despite the previous work, which only studies the \textit{AllTune} technique, we investigate the impact of paraphrases for various language model tuning techniques. Without contrastive learning, we can show that we can improve LM's robustness by paraphrase generation during training.

\noindent
{\bf Prompt Optimization \& Efficient Tuning}: Recent research proposes various techniques for prompt optimization and efficient tuning. We have used successful techniques from each of these areas. Appendix~\ref{related-works-extra:appendix} provides our detailed description of these recent techniques.
All of the recent techniques for prompt optimization and efficient tuning use the original input context provided within the dataset.

\noindent
{\bf Paraphrase Generation}:
Recent techniques encompass various approaches, including the use of copy mechanisms, Variational Autoencoders, Generative Adversarial Networks, and Reinforcement Learning techniques to generate diverse paraphrases~\cite{zhou-bhat-2021-paraphrase}. While previous studies have applied RL techniques for paraphrase generation, we propose the use of MML gradients instead of policy gradients to fine-tune our paraphrase model by the reward of a secondary classification task (see Appendix~\ref{related-works-extra:appendix} for more discussion).

\section{Conclusion}
We investigated the impact of incorporating input paraphrases while fine-tuning PLMs with recent efficient tuning techniques. Our results indicate that specific techniques, such as continuous and discrete prompt optimization methods like AutoPrompt or Soft-Prompt Tuning, benefit significantly from the inclusion of paraphrases. We also conducted extensive experiments to reduce noise in a distantly supervised paraphrase generator. Our ablation studies on fine-tuning the paraphrase generator demonstrate that policy gradient objectives lack robustness during training, while maximum marginal likelihood training remains a robust technique.

\section*{Ethics Statement}
Many language models show biases in their output due to the data used to train them \cite{liang2021towards}.   It is possible that even with few-shot language model tuning, we might continue to detect analogous biases in the downstream classification task, for instance, resulting in diminished classification accuracy for specific minority groups.  It is also possible that the additional data generated by the paraphrase model will exaggerate existing biases.  

\section*{Acknowledgements}
We gratefully acknowledge the Vector Institute, and the Digital Research Alliance of Canada (alliancecan.ca) for providing access to their GPU cluster, which streamlined our experiments. This research was funded by NSERC, Alberta Machine Intelligence Institute (Amii), and CIFAR. Alona Fyshe is a Canada CIFAR AI Chair.

\bibliography{my_bib, new_bib}

\begin{thebibliography}{71}
\expandafter\ifx\csname natexlab\endcsname\relax\def\natexlab#1{#1}\fi

\bibitem[{Abaskohi et~al.(2023)Abaskohi, Rothe, and
  Yaghoobzadeh}]{abaskohi-etal-2023-lm}
Amirhossein Abaskohi, Sascha Rothe, and Yadollah Yaghoobzadeh. 2023.
\newblock \href {https://doi.org/10.18653/v1/2023.acl-short.59} {{LM}-{CPPF}:
  Paraphrasing-guided data augmentation for contrastive prompt-based few-shot
  fine-tuning}.
\newblock In \emph{Proceedings of the 61st Annual Meeting of the Association
  for Computational Linguistics (Volume 2: Short Papers)}, pages 670--681,
  Toronto, Canada. Association for Computational Linguistics.

\bibitem[{Anil et~al.(2023)Anil, Dai, Firat, Johnson, Lepikhin, Passos,
  Shakeri, Taropa, Bailey, Chen, Chu, Clark, Shafey, Huang, Meier-Hellstern,
  Mishra, Moreira, Omernick, Robinson, Ruder, Tay, Xiao, Xu, Zhang, Abrego,
  Ahn, Austin, Barham, Botha, Bradbury, Brahma, Brooks, Catasta, Cheng, Cherry,
  Choquette-Choo, Chowdhery, Crepy, Dave, Dehghani, Dev, Devlin, Díaz, Du,
  Dyer, Feinberg, Feng, Fienber, Freitag, Garcia, Gehrmann, Gonzalez, Gur-Ari,
  Hand, Hashemi, Hou, Howland, Hu, Hui, Hurwitz, Isard, Ittycheriah, Jagielski,
  Jia, Kenealy, Krikun, Kudugunta, Lan, Lee, Lee, Li, Li, Li, Li, Li, Lim, Lin,
  Liu, Liu, Maggioni, Mahendru, Maynez, Misra, Moussalem, Nado, Nham, Ni,
  Nystrom, Parrish, Pellat, Polacek, Polozov, Pope, Qiao, Reif, Richter, Riley,
  Ros, Roy, Saeta, Samuel, Shelby, Slone, Smilkov, So, Sohn, Tokumine, Valter,
  Vasudevan, Vodrahalli, Wang, Wang, Wang, Wang, Wieting, Wu, Xu, Xu, Xue, Yin,
  Yu, Zhang, Zheng, Zheng, Zhou, Zhou, Petrov, and Wu}]{anil2023palm}
Rohan Anil, Andrew~M. Dai, Orhan Firat, Melvin Johnson, Dmitry Lepikhin,
  Alexandre Passos, Siamak Shakeri, Emanuel Taropa, Paige Bailey, Zhifeng Chen,
  Eric Chu, Jonathan~H. Clark, Laurent~El Shafey, Yanping Huang, Kathy
  Meier-Hellstern, Gaurav Mishra, Erica Moreira, Mark Omernick, Kevin Robinson,
  Sebastian Ruder, Yi~Tay, Kefan Xiao, Yuanzhong Xu, Yujing Zhang,
  Gustavo~Hernandez Abrego, Junwhan Ahn, Jacob Austin, Paul Barham, Jan Botha,
  James Bradbury, Siddhartha Brahma, Kevin Brooks, Michele Catasta, Yong Cheng,
  Colin Cherry, Christopher~A. Choquette-Choo, Aakanksha Chowdhery, Clément
  Crepy, Shachi Dave, Mostafa Dehghani, Sunipa Dev, Jacob Devlin, Mark Díaz,
  Nan Du, Ethan Dyer, Vlad Feinberg, Fangxiaoyu Feng, Vlad Fienber, Markus
  Freitag, Xavier Garcia, Sebastian Gehrmann, Lucas Gonzalez, Guy Gur-Ari,
  Steven Hand, Hadi Hashemi, Le~Hou, Joshua Howland, Andrea Hu, Jeffrey Hui,
  Jeremy Hurwitz, Michael Isard, Abe Ittycheriah, Matthew Jagielski, Wenhao
  Jia, Kathleen Kenealy, Maxim Krikun, Sneha Kudugunta, Chang Lan, Katherine
  Lee, Benjamin Lee, Eric Li, Music Li, Wei Li, YaGuang Li, Jian Li, Hyeontaek
  Lim, Hanzhao Lin, Zhongtao Liu, Frederick Liu, Marcello Maggioni, Aroma
  Mahendru, Joshua Maynez, Vedant Misra, Maysam Moussalem, Zachary Nado, John
  Nham, Eric Ni, Andrew Nystrom, Alicia Parrish, Marie Pellat, Martin Polacek,
  Alex Polozov, Reiner Pope, Siyuan Qiao, Emily Reif, Bryan Richter, Parker
  Riley, Alex~Castro Ros, Aurko Roy, Brennan Saeta, Rajkumar Samuel, Renee
  Shelby, Ambrose Slone, Daniel Smilkov, David~R. So, Daniel Sohn, Simon
  Tokumine, Dasha Valter, Vijay Vasudevan, Kiran Vodrahalli, Xuezhi Wang,
  Pidong Wang, Zirui Wang, Tao Wang, John Wieting, Yuhuai Wu, Kelvin Xu, Yunhan
  Xu, Linting Xue, Pengcheng Yin, Jiahui Yu, Qiao Zhang, Steven Zheng,
  Ce~Zheng, Weikang Zhou, Denny Zhou, Slav Petrov, and Yonghui Wu. 2023.
\newblock \href {http://arxiv.org/abs/2305.10403} {Palm 2 technical report}.

\bibitem[{Broughton(1995)}]{Broughton1995}
David Broughton. 1995.
\newblock \href {https://doi.org/10.1007/978-1-349-14907-0_2} {\emph{The
  assumptions and theory of public opinion polling}}, pages 15--33. Macmillan
  Education UK, London.

\bibitem[{Brown et~al.(2020{\natexlab{a}})Brown, Mann, Ryder, Subbiah, Kaplan,
  Dhariwal, Neelakantan, Shyam, Sastry, Askell, Agarwal, Herbert-Voss, Krueger,
  Henighan, Child, Ramesh, Ziegler, Wu, Winter, Hesse, Chen, Sigler, Litwin,
  Gray, Chess, Clark, Berner, McCandlish, Radford, Sutskever, and
  Amodei}]{brown2020language}
Tom~B. Brown, Benjamin Mann, Nick Ryder, Melanie Subbiah, Jared Kaplan,
  Prafulla Dhariwal, Arvind Neelakantan, Pranav Shyam, Girish Sastry, Amanda
  Askell, Sandhini Agarwal, Ariel Herbert-Voss, Gretchen Krueger, Tom Henighan,
  Rewon Child, Aditya Ramesh, Daniel~M. Ziegler, Jeffrey Wu, Clemens Winter,
  Christopher Hesse, Mark Chen, Eric Sigler, Mateusz Litwin, Scott Gray,
  Benjamin Chess, Jack Clark, Christopher Berner, Sam McCandlish, Alec Radford,
  Ilya Sutskever, and Dario Amodei. 2020{\natexlab{a}}.
\newblock \href {http://arxiv.org/abs/2005.14165} {Language models are few-shot
  learners}.

\bibitem[{Brown et~al.(2020{\natexlab{b}})Brown, Mann, Ryder, Subbiah, Kaplan,
  Dhariwal, Neelakantan, Shyam, Sastry, Askell, Agarwal, Herbert{-}Voss,
  Krueger, Henighan, Child, Ramesh, Ziegler, Wu, Winter, Hesse, Chen, Sigler,
  Litwin, Gray, Chess, Clark, Berner, McCandlish, Radford, Sutskever, and
  Amodei}]{DBLP:journals/corr/abs-2005-14165}
Tom~B. Brown, Benjamin Mann, Nick Ryder, Melanie Subbiah, Jared Kaplan,
  Prafulla Dhariwal, Arvind Neelakantan, Pranav Shyam, Girish Sastry, Amanda
  Askell, Sandhini Agarwal, Ariel Herbert{-}Voss, Gretchen Krueger, Tom
  Henighan, Rewon Child, Aditya Ramesh, Daniel~M. Ziegler, Jeffrey Wu, Clemens
  Winter, Christopher Hesse, Mark Chen, Eric Sigler, Mateusz Litwin, Scott
  Gray, Benjamin Chess, Jack Clark, Christopher Berner, Sam McCandlish, Alec
  Radford, Ilya Sutskever, and Dario Amodei. 2020{\natexlab{b}}.
\newblock \href {http://arxiv.org/abs/2005.14165} {Language models are few-shot
  learners}.
\newblock \emph{CoRR}, abs/2005.14165.

\bibitem[{Chen et~al.(2021)Chen, Tam, Raffel, Bansal, and
  Yang}]{DBLP:journals/corr/abs-2106-07499}
Jiaao Chen, Derek Tam, Colin Raffel, Mohit Bansal, and Diyi Yang. 2021.
\newblock \href {http://arxiv.org/abs/2106.07499} {An empirical survey of data
  augmentation for limited data learning in {NLP}}.
\newblock \emph{CoRR}, abs/2106.07499.

\bibitem[{Cui et~al.(2023)Cui, Li, Ding, Huang, Liu, and
  Sun}]{cui-etal-2023-decoder}
Ganqu Cui, Wentao Li, Ning Ding, Longtao Huang, Zhiyuan Liu, and Maosong Sun.
  2023.
\newblock \href {https://aclanthology.org/2023.acl-long.840} {Decoder tuning:
  Efficient language understanding as decoding}.
\newblock In \emph{Proceedings of the 61st Annual Meeting of the Association
  for Computational Linguistics (Volume 1: Long Papers)}, pages 15072--15087,
  Toronto, Canada. Association for Computational Linguistics.

\bibitem[{Deng et~al.(2022)Deng, Wang, Hsieh, Wang, Guo, Shu, Song, Xing, and
  Hu}]{deng-etal-2022-rlprompt}
Mingkai Deng, Jianyu Wang, Cheng-Ping Hsieh, Yihan Wang, Han Guo, Tianmin Shu,
  Meng Song, Eric Xing, and Zhiting Hu. 2022.
\newblock \href {https://aclanthology.org/2022.emnlp-main.222} {{RLP}rompt:
  Optimizing discrete text prompts with reinforcement learning}.
\newblock In \emph{Proceedings of the 2022 Conference on Empirical Methods in
  Natural Language Processing}, pages 3369--3391, Abu Dhabi, United Arab
  Emirates. Association for Computational Linguistics.

\bibitem[{Devlin et~al.(2019)Devlin, Chang, Lee, and
  Toutanova}]{devlin-etal-2019-bert}
Jacob Devlin, Ming-Wei Chang, Kenton Lee, and Kristina Toutanova. 2019.
\newblock \href {https://doi.org/10.18653/v1/N19-1423} {{BERT}: Pre-training of
  deep bidirectional transformers for language understanding}.
\newblock In \emph{Proceedings of the 2019 Conference of the North {A}merican
  Chapter of the Association for Computational Linguistics: Human Language
  Technologies, Volume 1 (Long and Short Papers)}, pages 4171--4186,
  Minneapolis, Minnesota. Association for Computational Linguistics.

\bibitem[{Donaldson(1978)}]{childBook}
Margaret~C. Donaldson. 1978.
\newblock \emph{Children's Minds}.

\bibitem[{Du and Ji(2019)}]{du-ji-2019-empirical}
Wanyu Du and Yangfeng Ji. 2019.
\newblock \href {https://doi.org/10.18653/v1/D19-1619} {An empirical comparison
  on imitation learning and reinforcement learning for paraphrase generation}.
\newblock In \emph{Proceedings of the 2019 Conference on Empirical Methods in
  Natural Language Processing and the 9th International Joint Conference on
  Natural Language Processing (EMNLP-IJCNLP)}, pages 6012--6018, Hong Kong,
  China. Association for Computational Linguistics.

\bibitem[{Feng et~al.(2021)Feng, Gangal, Wei, Chandar, Vosoughi, Mitamura, and
  Hovy}]{feng-etal-2021-survey}
Steven~Y. Feng, Varun Gangal, Jason Wei, Sarath Chandar, Soroush Vosoughi,
  Teruko Mitamura, and Eduard Hovy. 2021.
\newblock \href {https://doi.org/10.18653/v1/2021.findings-acl.84} {A survey of
  data augmentation approaches for {NLP}}.
\newblock In \emph{Findings of the Association for Computational Linguistics:
  ACL-IJCNLP 2021}, pages 968--988, Online. Association for Computational
  Linguistics.

\bibitem[{Gao et~al.(2021)Gao, Fisch, and Chen}]{gao-etal-2021-making}
Tianyu Gao, Adam Fisch, and Danqi Chen. 2021.
\newblock \href {https://doi.org/10.18653/v1/2021.acl-long.295} {Making
  pre-trained language models better few-shot learners}.
\newblock In \emph{Proceedings of the 59th Annual Meeting of the Association
  for Computational Linguistics and the 11th International Joint Conference on
  Natural Language Processing (Volume 1: Long Papers)}, pages 3816--3830,
  Online. Association for Computational Linguistics.

\bibitem[{Garg et~al.(2021)Garg, Prabhu, Misra, and
  Srinivasaraghavan}]{DBLP:journals/corr/abs-2103-12777}
Sonal Garg, Sumanth Prabhu, Hemant Misra, and G.~Srinivasaraghavan. 2021.
\newblock \href {http://arxiv.org/abs/2103.12777} {Unsupervised contextual
  paraphrase generation using lexical control and reinforcement learning}.
\newblock \emph{CoRR}, abs/2103.12777.

\bibitem[{Guo et~al.(2022)Guo, Tan, Liu, Xing, and
  Hu}]{guo-etal-2022-efficient}
Han Guo, Bowen Tan, Zhengzhong Liu, Eric Xing, and Zhiting Hu. 2022.
\newblock \href {https://aclanthology.org/2022.findings-emnlp.518} {Efficient
  (soft) {Q}-learning for text generation with limited good data}.
\newblock In \emph{Findings of the Association for Computational Linguistics:
  EMNLP 2022}, pages 6969--6991, Abu Dhabi, United Arab Emirates. Association
  for Computational Linguistics.

\bibitem[{Guo et~al.(2021)Guo, Tan, Liu, Xing, and
  Hu}]{https://doi.org/10.48550/arxiv.2106.07704}
Han Guo, Bowen Tan, Zhengzhong Liu, Eric~P. Xing, and Zhiting Hu. 2021.
\newblock \href {https://doi.org/10.48550/ARXIV.2106.07704} {Efficient (soft)
  q-learning for text generation with limited good data}.

\bibitem[{Guu et~al.(2017)Guu, Pasupat, Liu, and
  Liang}]{guu-etal-2017-language}
Kelvin Guu, Panupong Pasupat, Evan Liu, and Percy Liang. 2017.
\newblock \href {https://doi.org/10.18653/v1/P17-1097} {From language to
  programs: Bridging reinforcement learning and maximum marginal likelihood}.
\newblock In \emph{Proceedings of the 55th Annual Meeting of the Association
  for Computational Linguistics (Volume 1: Long Papers)}, pages 1051--1062,
  Vancouver, Canada. Association for Computational Linguistics.

\bibitem[{Hendrycks and Gimpel(2016)}]{DBLP:journals/corr/HendrycksG16}
Dan Hendrycks and Kevin Gimpel. 2016.
\newblock \href {http://arxiv.org/abs/1606.08415} {Bridging nonlinearities and
  stochastic regularizers with gaussian error linear units}.
\newblock \emph{CoRR}, abs/1606.08415.

\bibitem[{Holtzman et~al.(2020)Holtzman, Buys, Du, Forbes, and
  Choi}]{holtzman2020curious}
Ari Holtzman, Jan Buys, Li~Du, Maxwell Forbes, and Yejin Choi. 2020.
\newblock \href {http://arxiv.org/abs/1904.09751} {The curious case of neural
  text degeneration}.

\bibitem[{Houlsby et~al.(2019{\natexlab{a}})Houlsby, Giurgiu, Jastrzebski,
  Morrone, De~Laroussilhe, Gesmundo, Attariyan, and
  Gelly}]{pmlr-v97-houlsby19a}
Neil Houlsby, Andrei Giurgiu, Stanislaw Jastrzebski, Bruna Morrone, Quentin
  De~Laroussilhe, Andrea Gesmundo, Mona Attariyan, and Sylvain Gelly.
  2019{\natexlab{a}}.
\newblock \href {https://proceedings.mlr.press/v97/houlsby19a.html}
  {Parameter-efficient transfer learning for {NLP}}.
\newblock In \emph{Proceedings of the 36th International Conference on Machine
  Learning}, volume~97 of \emph{Proceedings of Machine Learning Research},
  pages 2790--2799. PMLR.

\bibitem[{Houlsby et~al.(2019{\natexlab{b}})Houlsby, Giurgiu, Jastrzebski,
  Morrone, de~Laroussilhe, Gesmundo, Attariyan, and
  Gelly}]{DBLP:journals/corr/abs-1902-00751}
Neil Houlsby, Andrei Giurgiu, Stanislaw Jastrzebski, Bruna Morrone, Quentin
  de~Laroussilhe, Andrea Gesmundo, Mona Attariyan, and Sylvain Gelly.
  2019{\natexlab{b}}.
\newblock \href {http://arxiv.org/abs/1902.00751} {Parameter-efficient transfer
  learning for {NLP}}.
\newblock \emph{CoRR}, abs/1902.00751.

\bibitem[{Hu et~al.(2021)Hu, Shen, Wallis, Allen{-}Zhu, Li, Wang, and
  Chen}]{DBLP:journals/corr/abs-2106-09685}
Edward~J. Hu, Yelong Shen, Phillip Wallis, Zeyuan Allen{-}Zhu, Yuanzhi Li,
  Shean Wang, and Weizhu Chen. 2021.
\newblock \href {http://arxiv.org/abs/2106.09685} {Lora: Low-rank adaptation of
  large language models}.
\newblock \emph{CoRR}, abs/2106.09685.

\bibitem[{Hu and Liu(2004)}]{10.1145/1014052.1014073}
Minqing Hu and Bing Liu. 2004.
\newblock \href {https://doi.org/10.1145/1014052.1014073} {Mining and
  summarizing customer reviews}.
\newblock In \emph{Proceedings of the Tenth ACM SIGKDD International Conference
  on Knowledge Discovery and Data Mining}, KDD '04, page 168–177, New York,
  NY, USA. Association for Computing Machinery.

\bibitem[{Izmailov et~al.(2019)Izmailov, Podoprikhin, Garipov, Vetrov, and
  Wilson}]{izmailov2019averaging}
Pavel Izmailov, Dmitrii Podoprikhin, Timur Garipov, Dmitry Vetrov, and
  Andrew~Gordon Wilson. 2019.
\newblock \href {http://arxiv.org/abs/1803.05407} {Averaging weights leads to
  wider optima and better generalization}.

\bibitem[{Kojima et~al.(2022)Kojima, Gu, Reid, Matsuo, and
  Iwasawa}]{NEURIPS2022_8bb0d291}
Takeshi Kojima, Shixiang~(Shane) Gu, Machel Reid, Yutaka Matsuo, and Yusuke
  Iwasawa. 2022.
\newblock \href
  {https://proceedings.neurips.cc/paper_files/paper/2022/file/8bb0d291acd4acf06ef112099c16f326-Paper-Conference.pdf}
  {Large language models are zero-shot reasoners}.
\newblock In \emph{Advances in Neural Information Processing Systems},
  volume~35, pages 22199--22213. Curran Associates, Inc.

\bibitem[{Lester et~al.(2021)Lester, Al-Rfou, and
  Constant}]{lester-etal-2021-power}
Brian Lester, Rami Al-Rfou, and Noah Constant. 2021.
\newblock \href {https://doi.org/10.18653/v1/2021.emnlp-main.243} {The power of
  scale for parameter-efficient prompt tuning}.
\newblock In \emph{Proceedings of the 2021 Conference on Empirical Methods in
  Natural Language Processing}, pages 3045--3059, Online and Punta Cana,
  Dominican Republic. Association for Computational Linguistics.

\bibitem[{Li and Liang(2021)}]{li-liang-2021-prefix}
Xiang~Lisa Li and Percy Liang. 2021.
\newblock \href {https://doi.org/10.18653/v1/2021.acl-long.353} {Prefix-tuning:
  Optimizing continuous prompts for generation}.
\newblock In \emph{Proceedings of the 59th Annual Meeting of the Association
  for Computational Linguistics and the 11th International Joint Conference on
  Natural Language Processing (Volume 1: Long Papers)}, pages 4582--4597,
  Online. Association for Computational Linguistics.

\bibitem[{Li et~al.(2023)Li, Lin, Zhang, Fu, Chen, Lou, and
  Chen}]{li-etal-2023-making}
Yifei Li, Zeqi Lin, Shizhuo Zhang, Qiang Fu, Bei Chen, Jian-Guang Lou, and
  Weizhu Chen. 2023.
\newblock \href {https://doi.org/10.18653/v1/2023.acl-long.291} {Making
  language models better reasoners with step-aware verifier}.
\newblock In \emph{Proceedings of the 61st Annual Meeting of the Association
  for Computational Linguistics (Volume 1: Long Papers)}, pages 5315--5333,
  Toronto, Canada. Association for Computational Linguistics.

\bibitem[{Li et~al.(2018)Li, Jiang, Shang, and Li}]{li-etal-2018-paraphrase}
Zichao Li, Xin Jiang, Lifeng Shang, and Hang Li. 2018.
\newblock \href {https://doi.org/10.18653/v1/D18-1421} {Paraphrase generation
  with deep reinforcement learning}.
\newblock In \emph{Proceedings of the 2018 Conference on Empirical Methods in
  Natural Language Processing}, pages 3865--3878, Brussels, Belgium.
  Association for Computational Linguistics.

\bibitem[{Liang et~al.(2018)Liang, Norouzi, Berant, Le, and
  Lao}]{NEURIPS2018_f4e369c0}
Chen Liang, Mohammad Norouzi, Jonathan Berant, Quoc~V Le, and Ni~Lao. 2018.
\newblock \href
  {https://proceedings.neurips.cc/paper_files/paper/2018/file/f4e369c0a468d3aeeda0593ba90b5e55-Paper.pdf}
  {Memory augmented policy optimization for program synthesis and semantic
  parsing}.
\newblock In \emph{Advances in Neural Information Processing Systems},
  volume~31. Curran Associates, Inc.

\bibitem[{Liang et~al.(2021)Liang, Wu, Morency, and
  Salakhutdinov}]{liang2021towards}
Paul~Pu Liang, Chiyu Wu, Louis-Philippe Morency, and Ruslan Salakhutdinov.
  2021.
\newblock {Towards Understanding and Mitigating Social Biases in Language
  Models}.
\newblock In \emph{International Conference on Machine Learning}, pages
  6565--6576. PMLR.

\bibitem[{Lin(2004)}]{lin-2004-rouge}
Chin-Yew Lin. 2004.
\newblock \href {https://aclanthology.org/W04-1013} {{ROUGE}: A package for
  automatic evaluation of summaries}.
\newblock In \emph{Text Summarization Branches Out}, pages 74--81, Barcelona,
  Spain. Association for Computational Linguistics.

\bibitem[{Lin et~al.(2020)Lin, Madotto, and Fung}]{lin-etal-2020-exploring}
Zhaojiang Lin, Andrea Madotto, and Pascale Fung. 2020.
\newblock \href {https://doi.org/10.18653/v1/2020.findings-emnlp.41} {Exploring
  versatile generative language model via parameter-efficient transfer
  learning}.
\newblock In \emph{Findings of the Association for Computational Linguistics:
  EMNLP 2020}, pages 441--459, Online. Association for Computational
  Linguistics.

\bibitem[{Liu et~al.(2020)Liu, Yang, Xiong, Zhang, Meng, Hu, Xu, and
  Chen}]{liu-etal-2020-learning}
Mingtong Liu, Erguang Yang, Deyi Xiong, Yujie Zhang, Yao Meng, Changjian Hu,
  Jinan Xu, and Yufeng Chen. 2020.
\newblock \href {https://doi.org/10.18653/v1/2020.coling-main.209} {A
  learning-exploring method to generate diverse paraphrases with
  multi-objective deep reinforcement learning}.
\newblock In \emph{Proceedings of the 28th International Conference on
  Computational Linguistics}, pages 2310--2321, Barcelona, Spain (Online).
  International Committee on Computational Linguistics.

\bibitem[{Liu et~al.(2021{\natexlab{a}})Liu, Yuan, Fu, Jiang, Hayashi, and
  Neubig}]{DBLP:journals/corr/abs-2107-13586}
Pengfei Liu, Weizhe Yuan, Jinlan Fu, Zhengbao Jiang, Hiroaki Hayashi, and
  Graham Neubig. 2021{\natexlab{a}}.
\newblock \href {http://arxiv.org/abs/2107.13586} {Pre-train, prompt, and
  predict: {A} systematic survey of prompting methods in natural language
  processing}.
\newblock \emph{CoRR}, abs/2107.13586.

\bibitem[{Liu et~al.(2021{\natexlab{b}})Liu, Yuan, Fu, Jiang, Hayashi, and
  Neubig}]{liu2021pretrain}
Pengfei Liu, Weizhe Yuan, Jinlan Fu, Zhengbao Jiang, Hiroaki Hayashi, and
  Graham Neubig. 2021{\natexlab{b}}.
\newblock \href {http://arxiv.org/abs/2107.13586} {Pre-train, prompt, and
  predict: A systematic survey of prompting methods in natural language
  processing}.

\bibitem[{Liu et~al.(2022)Liu, Ji, Fu, Tam, Du, Yang, and
  Tang}]{liu-etal-2022-p}
Xiao Liu, Kaixuan Ji, Yicheng Fu, Weng Tam, Zhengxiao Du, Zhilin Yang, and Jie
  Tang. 2022.
\newblock \href {https://doi.org/10.18653/v1/2022.acl-short.8} {{P}-tuning:
  Prompt tuning can be comparable to fine-tuning across scales and tasks}.
\newblock In \emph{Proceedings of the 60th Annual Meeting of the Association
  for Computational Linguistics (Volume 2: Short Papers)}, pages 61--68,
  Dublin, Ireland. Association for Computational Linguistics.

\bibitem[{Liu et~al.(2019)Liu, Ott, Goyal, Du, Joshi, Chen, Levy, Lewis,
  Zettlemoyer, and Stoyanov}]{DBLP:journals/corr/abs-1907-11692}
Yinhan Liu, Myle Ott, Naman Goyal, Jingfei Du, Mandar Joshi, Danqi Chen, Omer
  Levy, Mike Lewis, Luke Zettlemoyer, and Veselin Stoyanov. 2019.
\newblock \href {http://arxiv.org/abs/1907.11692} {Roberta: {A} robustly
  optimized {BERT} pretraining approach}.
\newblock \emph{CoRR}, abs/1907.11692.

\bibitem[{Loshchilov and Hutter(2017)}]{DBLP:journals/corr/abs-1711-05101}
Ilya Loshchilov and Frank Hutter. 2017.
\newblock \href {http://arxiv.org/abs/1711.05101} {Fixing weight decay
  regularization in adam}.
\newblock \emph{CoRR}, abs/1711.05101.

\bibitem[{Mnih et~al.(2016)Mnih, Badia, Mirza, Graves, Lillicrap, Harley,
  Silver, and Kavukcuoglu}]{DBLP:journals/corr/MnihBMGLHSK16}
Volodymyr Mnih, Adri{\`{a}}~Puigdom{\`{e}}nech Badia, Mehdi Mirza, Alex Graves,
  Timothy~P. Lillicrap, Tim Harley, David Silver, and Koray Kavukcuoglu. 2016.
\newblock \href {http://arxiv.org/abs/1602.01783} {Asynchronous methods for
  deep reinforcement learning}.
\newblock \emph{CoRR}, abs/1602.01783.

\bibitem[{Najafi and Fyshe(2023)}]{najafi-fyshe-2023-weakly}
Saeed Najafi and Alona Fyshe. 2023.
\newblock \href {https://aclanthology.org/2023.eacl-main.224}
  {Weakly-supervised questions for zero-shot relation extraction}.
\newblock In \emph{Proceedings of the 17th Conference of the European Chapter
  of the Association for Computational Linguistics}, pages 3075--3087,
  Dubrovnik, Croatia. Association for Computational Linguistics.

\bibitem[{Pang and Lee(2005)}]{pang-lee-2005-seeing}
Bo~Pang and Lillian Lee. 2005.
\newblock \href {https://doi.org/10.3115/1219840.1219855} {Seeing stars:
  Exploiting class relationships for sentiment categorization with respect to
  rating scales}.
\newblock In \emph{Proceedings of the 43rd Annual Meeting of the Association
  for Computational Linguistics ({ACL}{'}05)}, pages 115--124, Ann Arbor,
  Michigan. Association for Computational Linguistics.

\bibitem[{Petroni et~al.(2019)Petroni, Rockt{\"a}schel, Riedel, Lewis, Bakhtin,
  Wu, and Miller}]{petroni-etal-2019-language}
Fabio Petroni, Tim Rockt{\"a}schel, Sebastian Riedel, Patrick Lewis, Anton
  Bakhtin, Yuxiang Wu, and Alexander Miller. 2019.
\newblock \href {https://doi.org/10.18653/v1/D19-1250} {Language models as
  knowledge bases?}
\newblock In \emph{Proceedings of the 2019 Conference on Empirical Methods in
  Natural Language Processing and the 9th International Joint Conference on
  Natural Language Processing (EMNLP-IJCNLP)}, pages 2463--2473, Hong Kong,
  China. Association for Computational Linguistics.

\bibitem[{Qian et~al.(2019)Qian, Qiu, Zhang, Jiang, and
  Yu}]{qian-etal-2019-exploring}
Lihua Qian, Lin Qiu, Weinan Zhang, Xin Jiang, and Yong Yu. 2019.
\newblock \href {https://doi.org/10.18653/v1/D19-1313} {Exploring diverse
  expressions for paraphrase generation}.
\newblock In \emph{Proceedings of the 2019 Conference on Empirical Methods in
  Natural Language Processing and the 9th International Joint Conference on
  Natural Language Processing (EMNLP-IJCNLP)}, pages 3173--3182, Hong Kong,
  China. Association for Computational Linguistics.

\bibitem[{Radford et~al.(2019{\natexlab{a}})Radford, Wu, Child, Luan, Amodei,
  and Sutskever}]{Radford2019LanguageMA}
Alec Radford, Jeff Wu, Rewon Child, David Luan, Dario Amodei, and Ilya
  Sutskever. 2019{\natexlab{a}}.
\newblock \href {https://api.semanticscholar.org/CorpusID:160025533} {Language
  models are unsupervised multitask learners}.

\bibitem[{Radford et~al.(2019{\natexlab{b}})Radford, Wu, Child, Luan, Amodei,
  Sutskever et~al.}]{radford2019language}
Alec Radford, Jeffrey Wu, Rewon Child, David Luan, Dario Amodei, Ilya
  Sutskever, et~al. 2019{\natexlab{b}}.
\newblock Language models are unsupervised multitask learners.
\newblock \emph{OpenAI blog}, 1(8):9.

\bibitem[{Raffel et~al.(2019)Raffel, Shazeer, Roberts, Lee, Narang, Matena,
  Zhou, Li, and Liu}]{DBLP:journals/corr/abs-1910-10683}
Colin Raffel, Noam Shazeer, Adam Roberts, Katherine Lee, Sharan Narang, Michael
  Matena, Yanqi Zhou, Wei Li, and Peter~J. Liu. 2019.
\newblock \href {http://arxiv.org/abs/1910.10683} {Exploring the limits of
  transfer learning with a unified text-to-text transformer}.
\newblock \emph{CoRR}, abs/1910.10683.

\bibitem[{Reddi et~al.(2019)Reddi, Kale, and
  Kumar}]{DBLP:journals/corr/abs-1904-09237}
Sashank~J. Reddi, Satyen Kale, and Sanjiv Kumar. 2019.
\newblock \href {http://arxiv.org/abs/1904.09237} {On the convergence of adam
  and beyond}.
\newblock \emph{CoRR}, abs/1904.09237.

\bibitem[{Rennie et~al.(2016)Rennie, Marcheret, Mroueh, Ross, and
  Goel}]{DBLP:journals/corr/RennieMMRG16}
Steven~J. Rennie, Etienne Marcheret, Youssef Mroueh, Jerret Ross, and Vaibhava
  Goel. 2016.
\newblock \href {http://arxiv.org/abs/1612.00563} {Self-critical sequence
  training for image captioning}.
\newblock \emph{CoRR}, abs/1612.00563.

\bibitem[{Sancheti et~al.(2022)Sancheti, Srinivasan, and
  Rudinger}]{Sancheti_Srinivasan_Rudinger_2022}
Abhilasha Sancheti, Balaji~Vasan Srinivasan, and Rachel Rudinger. 2022.
\newblock \href {https://doi.org/10.1609/aaai.v36i10.21376} {Entailment
  relation aware paraphrase generation}.
\newblock \emph{Proceedings of the AAAI Conference on Artificial Intelligence},
  36(10):11258--11266.

\bibitem[{Shi et~al.(2022)Shi, Han, Gonen, Holtzman, Tsvetkov, and
  Zettlemoyer}]{shi2022human}
Weijia Shi, Xiaochuang Han, Hila Gonen, Ari Holtzman, Yulia Tsvetkov, and Luke
  Zettlemoyer. 2022.
\newblock \href {http://arxiv.org/abs/2212.10539} {Toward human readable prompt
  tuning: Kubrick's the shining is a good movie, and a good prompt too?}

\bibitem[{Shin et~al.(2020)Shin, Razeghi, Logan~IV, Wallace, and
  Singh}]{shin-etal-2020-autoprompt}
Taylor Shin, Yasaman Razeghi, Robert~L. Logan~IV, Eric Wallace, and Sameer
  Singh. 2020.
\newblock \href {https://doi.org/10.18653/v1/2020.emnlp-main.346}
  {{A}uto{P}rompt: {E}liciting {K}nowledge from {L}anguage {M}odels with
  {A}utomatically {G}enerated {P}rompts}.
\newblock In \emph{Proceedings of the 2020 Conference on Empirical Methods in
  Natural Language Processing (EMNLP)}, pages 4222--4235, Online. Association
  for Computational Linguistics.

\bibitem[{Siddique et~al.(2020)Siddique, Oymak, and
  Hristidis}]{10.1145/3394486.3403231}
A.~B. Siddique, Samet Oymak, and Vagelis Hristidis. 2020.
\newblock \href {https://doi.org/10.1145/3394486.3403231} {Unsupervised
  paraphrasing via deep reinforcement learning}.
\newblock In \emph{Proceedings of the 26th ACM SIGKDD International Conference
  on Knowledge Discovery \& Data Mining}, KDD '20, page 1800–1809, New York,
  NY, USA. Association for Computing Machinery.

\bibitem[{Socher et~al.(2013)Socher, Perelygin, Wu, Chuang, Manning, Ng, and
  Potts}]{socher-etal-2013-recursive}
Richard Socher, Alex Perelygin, Jean Wu, Jason Chuang, Christopher~D. Manning,
  Andrew Ng, and Christopher Potts. 2013.
\newblock \href {https://aclanthology.org/D13-1170} {Recursive deep models for
  semantic compositionality over a sentiment treebank}.
\newblock In \emph{Proceedings of the 2013 Conference on Empirical Methods in
  Natural Language Processing}, pages 1631--1642, Seattle, Washington, USA.
  Association for Computational Linguistics.

\bibitem[{Sun et~al.(2022)Sun, Shao, Qian, Huang, and
  Qiu}]{DBLP:journals/corr/abs-2201-03514}
Tianxiang Sun, Yunfan Shao, Hong Qian, Xuanjing Huang, and Xipeng Qiu. 2022.
\newblock \href {http://arxiv.org/abs/2201.03514} {Black-box tuning for
  language-model-as-a-service}.
\newblock \emph{CoRR}, abs/2201.03514.

\bibitem[{Sutton et~al.(1999)Sutton, McAllester, Singh, and
  Mansour}]{10.5555/3009657.3009806}
Richard~S. Sutton, David McAllester, Satinder Singh, and Yishay Mansour. 1999.
\newblock Policy gradient methods for reinforcement learning with function
  approximation.
\newblock In \emph{Proceedings of the 12th International Conference on Neural
  Information Processing Systems}, NIPS'99, page 1057–1063, Cambridge, MA,
  USA. MIT Press.

\bibitem[{Touvron et~al.(2023)Touvron, Martin, Stone, Albert, Almahairi,
  Babaei, Bashlykov, Batra, Bhargava, Bhosale, Bikel, Blecher, Ferrer, Chen,
  Cucurull, Esiobu, Fernandes, Fu, Fu, Fuller, Gao, Goswami, Goyal, Hartshorn,
  Hosseini, Hou, Inan, Kardas, Kerkez, Khabsa, Kloumann, Korenev, Koura,
  Lachaux, Lavril, Lee, Liskovich, Lu, Mao, Martinet, Mihaylov, Mishra,
  Molybog, Nie, Poulton, Reizenstein, Rungta, Saladi, Schelten, Silva, Smith,
  Subramanian, Tan, Tang, Taylor, Williams, Kuan, Xu, Yan, Zarov, Zhang, Fan,
  Kambadur, Narang, Rodriguez, Stojnic, Edunov, and Scialom}]{touvron2023llama}
Hugo Touvron, Louis Martin, Kevin Stone, Peter Albert, Amjad Almahairi, Yasmine
  Babaei, Nikolay Bashlykov, Soumya Batra, Prajjwal Bhargava, Shruti Bhosale,
  Dan Bikel, Lukas Blecher, Cristian~Canton Ferrer, Moya Chen, Guillem
  Cucurull, David Esiobu, Jude Fernandes, Jeremy Fu, Wenyin Fu, Brian Fuller,
  Cynthia Gao, Vedanuj Goswami, Naman Goyal, Anthony Hartshorn, Saghar
  Hosseini, Rui Hou, Hakan Inan, Marcin Kardas, Viktor Kerkez, Madian Khabsa,
  Isabel Kloumann, Artem Korenev, Punit~Singh Koura, Marie-Anne Lachaux,
  Thibaut Lavril, Jenya Lee, Diana Liskovich, Yinghai Lu, Yuning Mao, Xavier
  Martinet, Todor Mihaylov, Pushkar Mishra, Igor Molybog, Yixin Nie, Andrew
  Poulton, Jeremy Reizenstein, Rashi Rungta, Kalyan Saladi, Alan Schelten, Ruan
  Silva, Eric~Michael Smith, Ranjan Subramanian, Xiaoqing~Ellen Tan, Binh Tang,
  Ross Taylor, Adina Williams, Jian~Xiang Kuan, Puxin Xu, Zheng Yan, Iliyan
  Zarov, Yuchen Zhang, Angela Fan, Melanie Kambadur, Sharan Narang, Aurelien
  Rodriguez, Robert Stojnic, Sergey Edunov, and Thomas Scialom. 2023.
\newblock \href {http://arxiv.org/abs/2307.09288} {Llama 2: Open foundation and
  fine-tuned chat models}.

\bibitem[{Valipour et~al.(2023)Valipour, Rezagholizadeh, Kobyzev, and
  Ghodsi}]{valipour-etal-2023-dyLoRA}
Mojtaba Valipour, Mehdi Rezagholizadeh, Ivan Kobyzev, and Ali Ghodsi. 2023.
\newblock \href {https://aclanthology.org/2023.eacl-main.239} {{D}y{L}o{RA}:
  Parameter-efficient tuning of pre-trained models using dynamic search-free
  low-rank adaptation}.
\newblock In \emph{Proceedings of the 17th Conference of the European Chapter
  of the Association for Computational Linguistics}, pages 3274--3287,
  Dubrovnik, Croatia. Association for Computational Linguistics.

\bibitem[{Vijayakumar et~al.(2018)Vijayakumar, Cogswell, Selvaraju, Sun, Lee,
  Crandall, and
  Batra}]{Vijayakumar_Cogswell_Selvaraju_Sun_Lee_Crandall_Batra_2018}
Ashwin Vijayakumar, Michael Cogswell, Ramprasaath Selvaraju, Qing Sun, Stefan
  Lee, David Crandall, and Dhruv Batra. 2018.
\newblock \href {https://doi.org/10.1609/aaai.v32i1.12340} {Diverse beam search
  for improved description of complex scenes}.
\newblock \emph{Proceedings of the AAAI Conference on Artificial Intelligence},
  32(1).

\bibitem[{Vladimir~Vorobev(2023)}]{chatgpt_paraphrases_dataset}
Maxim~Kuznetsov Vladimir~Vorobev. 2023.
\newblock Chatgpt paraphrases dataset.

\bibitem[{Voorhees and Tice(2000)}]{10.1145/345508.345577}
Ellen~M. Voorhees and Dawn~M. Tice. 2000.
\newblock \href {https://doi.org/10.1145/345508.345577} {Building a question
  answering test collection}.
\newblock In \emph{Proceedings of the 23rd Annual International ACM SIGIR
  Conference on Research and Development in Information Retrieval}, SIGIR '00,
  page 200–207, New York, NY, USA. Association for Computing Machinery.

\bibitem[{Wei et~al.(2022)Wei, Wang, Schuurmans, Bosma, Chi, Le, and
  Zhou}]{DBLP:journals/corr/abs-2201-11903}
Jason Wei, Xuezhi Wang, Dale Schuurmans, Maarten Bosma, Ed~H. Chi, Quoc Le, and
  Denny Zhou. 2022.
\newblock \href {http://arxiv.org/abs/2201.11903} {Chain of thought prompting
  elicits reasoning in large language models}.
\newblock \emph{CoRR}, abs/2201.11903.

\bibitem[{Wei and Zou(2019)}]{wei-zou-2019-eda}
Jason Wei and Kai Zou. 2019.
\newblock \href {https://doi.org/10.18653/v1/D19-1670} {{EDA}: Easy data
  augmentation techniques for boosting performance on text classification
  tasks}.
\newblock In \emph{Proceedings of the 2019 Conference on Empirical Methods in
  Natural Language Processing and the 9th International Joint Conference on
  Natural Language Processing (EMNLP-IJCNLP)}, pages 6382--6388, Hong Kong,
  China. Association for Computational Linguistics.

\bibitem[{Zhang et~al.(2023)Zhang, Chen, Bukharin, He, Cheng, Chen, and
  Zhao}]{zhang2023adaptive}
Qingru Zhang, Minshuo Chen, Alexander Bukharin, Pengcheng He, Yu~Cheng, Weizhu
  Chen, and Tuo Zhao. 2023.
\newblock \href {http://arxiv.org/abs/2303.10512} {Adaptive budget allocation
  for parameter-efficient fine-tuning}.

\bibitem[{Zhang et~al.(2022{\natexlab{a}})Zhang, Roller, Goyal, Artetxe, Chen,
  Chen, Dewan, Diab, Li, Lin, Mihaylov, Ott, Shleifer, Shuster, Simig, Koura,
  Sridhar, Wang, and Zettlemoyer}]{zhang2022opt}
Susan Zhang, Stephen Roller, Naman Goyal, Mikel Artetxe, Moya Chen, Shuohui
  Chen, Christopher Dewan, Mona Diab, Xian Li, Xi~Victoria Lin, Todor Mihaylov,
  Myle Ott, Sam Shleifer, Kurt Shuster, Daniel Simig, Punit~Singh Koura, Anjali
  Sridhar, Tianlu Wang, and Luke Zettlemoyer. 2022{\natexlab{a}}.
\newblock \href {http://arxiv.org/abs/2205.01068} {Opt: Open pre-trained
  transformer language models}.

\bibitem[{Zhang et~al.(2022{\natexlab{b}})Zhang, Wang, Zhou, Schuurmans, and
  Gonzalez}]{zhang2022tempera}
Tianjun Zhang, Xuezhi Wang, Denny Zhou, Dale Schuurmans, and Joseph~E.
  Gonzalez. 2022{\natexlab{b}}.
\newblock \href {http://arxiv.org/abs/2211.11890} {Tempera: Test-time prompting
  via reinforcement learning}.

\bibitem[{Zhang* et~al.(2020)Zhang*, Kishore*, Wu*, Weinberger, and
  Artzi}]{Zhang*2020BERTScore:}
Tianyi Zhang*, Varsha Kishore*, Felix Wu*, Kilian~Q. Weinberger, and Yoav
  Artzi. 2020.
\newblock \href {https://openreview.net/forum?id=SkeHuCVFDr} {Bertscore:
  Evaluating text generation with bert}.
\newblock In \emph{International Conference on Learning Representations}.

\bibitem[{Zhang et~al.(2015)Zhang, Zhao, and
  LeCun}]{DBLP:journals/corr/ZhangZL15}
Xiang Zhang, Junbo~Jake Zhao, and Yann LeCun. 2015.
\newblock \href {http://arxiv.org/abs/1509.01626} {Character-level
  convolutional networks for text classification}.
\newblock \emph{CoRR}, abs/1509.01626.

\bibitem[{Zhang et~al.(2022{\natexlab{c}})Zhang, Zhang, Li, and
  Smola}]{zhang2022automatic}
Zhuosheng Zhang, Aston Zhang, Mu~Li, and Alex Smola. 2022{\natexlab{c}}.
\newblock \href {http://arxiv.org/abs/2210.03493} {Automatic chain of thought
  prompting in large language models}.

\bibitem[{Zhou and Bhat(2021)}]{zhou-bhat-2021-paraphrase}
Jianing Zhou and Suma Bhat. 2021.
\newblock \href {https://doi.org/10.18653/v1/2021.emnlp-main.414} {Paraphrase
  generation: A survey of the state of the art}.
\newblock In \emph{Proceedings of the 2021 Conference on Empirical Methods in
  Natural Language Processing}, pages 5075--5086, Online and Punta Cana,
  Dominican Republic. Association for Computational Linguistics.

\bibitem[{Zhou et~al.(2023)Zhou, Muresanu, Han, Paster, Pitis, Chan, and
  Ba}]{zhou2023large}
Yongchao Zhou, Andrei~Ioan Muresanu, Ziwen Han, Keiran Paster, Silviu Pitis,
  Harris Chan, and Jimmy Ba. 2023.
\newblock \href {http://arxiv.org/abs/2211.01910} {Large language models are
  human-level prompt engineers}.

\end{thebibliography}

\appendix

\section{Baseline LM Tuning Techniques}
\label{lm-tuning-techniques:appendix}

\textbf{Gradient-Search (\textit{\textit{\textit{\textit{GS}}}}):}
The \textit{\textit{\textit{\textit{GS}}}} technique is based on the recent AutoPrompt~\cite{shin-etal-2020-autoprompt} method, which optimizes task instructions without updating any parameters in the model. The search process begins in the vocabulary space, optimizing the change in label log-likelihood when replacing token $p_{i}$ in the task instruction with another token $v$ from the vocabulary set. In our implementation, each search iteration randomly selects one mini-batch of training examples and then randomly selects a token from the task instruction to update. The top $k$ candidate tokens are determined based on the approximate change in label log-likelihood: $Top_v \; \{w^{T}_{v} . \nabla_{w_{p_i}} \log P_{\text{lm}}(y|p,x)\}$, where $w_v$ is the embedding vector of a candidate token $v$. The resulting $k$ new task instructions are evaluated again using label log-likelihood on the same training examples\footnote{The original AutoPrompt evaluates the new candidate instructions on another training mini-batch. For fewshot classification, we re-use the drawn training mini-batch to evaluate the complete new candidate instructions.}, and the top-performing instruction is retained for the next search iteration. Prompt optimization always uses the original input $x$ when searching new task prompts~\cite{shin-etal-2020-autoprompt, deng-etal-2022-rlprompt}. In our work, we investigate the impact of incorporating paraphrases of $x$ during search.

\textbf{Input-Finetuning (\textit{InTune}):} As a straightforward and efficient tuning technique, we compare to updating only the input embedding table in the transformer architecture. This method requires gradient computation similar to All-Finetuning (\textit{\textit{AllTune}}) as well as the \textit{\textit{\textit{GS}}} method.

\textbf{LM-Head-Finetuning (\textit{HTune}):} The transformer-based pre-trained language models consist of a language modeling head, which maps the hidden vectors to the token logit for each token in the vocabulary. For the \textit{\textit{HTune}} technique, we solely update the parameters of the language modeling head.

\textbf{Classifier-Finetuning (\textit{\textit{ClsTune}}):} In \textit{\textit{ClsTune}}, we first create a feature representation $h(x)$ for the input text $x$ using average pooling of the final hidden vectors in the last layer of the language model. Here, we assume that the language model (feature extractor) remains fixed, and we then construct a two-layer feedforward network with the $gelu$ activation function~\cite{DBLP:journals/corr/HendrycksG16} as a classification module on top of the language model.

\textbf{Softprompt-Tuning (\textit{\textit{SpTune}}):} In \textit{\textit{SpTune}} \cite{lester-etal-2021-power}, $L$ prompt tokens are prepended to the task instruction. These $L$ tokens are associated with $L$ dedicated prompt embedding vectors, extending the sequence of vectors derived from the task instruction and input text with an additional $L$ trainable feature vectors. During training, the original embedding table of the transformer model remains fixed, while a new prompt embedding table is trained by backpropagating the label log-likelihood into the prompt embedding table. In contrast to \textit{\textit{InTune}}, here the prompt vectors do not need to map to vocabulary words.

\textbf{Low-Rank Adaptation (\textit{\textit{LoRA}}):} \textit{\textit{LoRA}} is one of the latest efficient-tuning techniques specifically designed for PLMs~\cite{DBLP:journals/corr/abs-2106-09685}. It learns low-rank adaptation matrices for the query and value weight matrices within the transformer model. For a pre-trained weight matrix $W_q \in \mathbb{R}^{d \times k}$, \textit{\textit{LoRA}} learns the necessary adaptation (i.e., modification) of the weight matrix for a downstream task through a low-rank decomposition, expressed as $W_q + \triangle W_q \approx W_q + BA$. Here, $B \in \mathbb{R}^{d \times r}$, $A \in \mathbb{R}^{r \times k}$, and the rank $r \le min(d, k)$. The adaptation matrices $A$ and $B$ are the only parameters subject to training, while the original matrix $W_q$ does not receive any gradient updates. Studies have shown that \textit{\textit{LoRA}} performs on par with, or better than, \textit{\textit{AllTune}} across various PLMs~\cite{DBLP:journals/corr/abs-2106-09685}.
\\
All language model tuning techniques we have discussed will use the same input format. For example in the sentiment classification task, we use the following format:

{\small\textit{``<s> \{instruction\} \{text\} . It was <mask> . </s>''}.}

Except for \textit{\textit{ClsTune}}, all of our tuning techniques maximize the probability of the correct label token in place of the <mask> token. In contrast, \textit{\textit{ClsTune}} takes the formatted input and classifies it into one of the predefined class labels.

\section{Few-shot Paraphrase Fine-Tuning (Further Results)}
\label{training-paraphrase-extra:appendix}

\begin{table*}
\centering
\caption{The accuracy of the best performing validation checkpoint in the 128-shot SST2 classification task trained with the off-policy learning technique.}
\begin{tabular}{c | c c c | c}
\hline
& \multicolumn{3}{c|}{Off-Policy} & AVG \\
\small Learn Tech & \small{Top-P} & \small{Beam} & \small{Mixed} & \\
\hline
\small No Tuning & \small67.5 & \small67.5 & \small67.5 & \small67.5 \\
\hline
\small PG & \small68.6 \tiny (1.8 $\mid$ 67.5) & \small68.4 \tiny (1.6 $\mid$ 67.4) & \small69.1 \tiny (1.6 $\mid$ 67.5) & \small68.7 \tiny (1.7 $\mid$ 67.5) \\
\small PG-Z & \small 68.8 \tiny (1.7 $\mid$ 67.5) & \small 68.7 \tiny (1.1 $\mid$ 67.4) & \small 68.0 \tiny (2.1 $\mid$ 67.2) & \small 68.5 \tiny (1.6 $\mid$ 67.4) \\
\hline
\small MML & \small\textbf{69.2} \tiny (2.8 $\mid$ 68.0) & \small\textbf{70.1} \tiny (2.4 $\mid$ 68.6) & \small\textbf{70.1} \tiny (3.3 $\mid$ 68.4) & \small\textbf{69.8} \tiny (2.7 $\mid$ 68.3)\\
\small MML-Z & \small\textbf{69.2} \tiny (2.5 $\mid$ 68.3) & \small 69.7 \tiny (3.5 $\mid$ 68.6) & \small \textbf{70.1} \tiny (2.8 $\mid$ 68.6) & \small 69.7 \tiny (2.9 $\mid$ 68.5) \\
\hline
\end{tabular}
\label{maximum-curves-off}
\end{table*}

This section provides additional results that compare our training objectives for fine-tuning the paraphrase generator using the feedback from the downstream language model.

The off-policy learning technique improves performance when using basic rewards (i.e., 69.1\% compared to 67.9\% with mixed decoding). However, the combined effect of off-policy learning and reward normalization decreases performance. With mixed decoding, `PG-Z' yields an accuracy of 71.2\% in on-policy learning compared to an accuracy of 68.0\% with off-policy learning. The `AVG' column in Table~\ref{maximum-curves-off} further verifies this conclusion that reward normalization is not improving the final performance while training the model with off-policy learning. We hypothesize that with the off-policy learning technique, the normalized rewards should be re-weighted properly if the sampled paraphrases are from the fixed paraphrase model.

\section{Example Paraphrases}
\label{example-paraphrases:appendix}
In Table~\ref{example-one-paraphrase-without-riff} and Table~\ref{example-two-paraphrase-without-riff}, we present example paraphrases from the AGNews test dataset generated by our pre-trained paraphrase model. We selected the AGNews dataset for its suitability in paraphrasing longer texts or short paragraphs. Subsequently, in Table~\ref{example-one-paraphrase-with-riff} and Table~\ref{example-two-paraphrase-with-riff}, we display the generated paraphrases after fine-tuning the paraphrase model with the RIFF objective. Fewer hallucinations can be observed in the new paraphrases, which are highlighted in red.

\begin{table*}
\small
\centering
\caption{Generated top eight paraphrases using diverse beam search {\textbf{with the pre-trained paraphrase model (distilled from GPT3.5-turbo)}} for the following input sentence from the AGNews test dataset: \\
``\textit{Gardner Loses Quest for Repeat Wrestling Gold US heavyweight Rulon Gardner lost his Olympic title Wednesday after being beaten in the semi-final stage of the 120kg Greco-Roman wrestling event by Georgiy Tsurtsumia of Kazakhstan.}'' Contradicting information or hallucinations are highlighted in {\color{red} red}.}

\begin{tabular}{ p{0.1\linewidth} | p{0.87\linewidth} }
\hline
Paraphrase & Text \\
\hline
P1 & ``\textit{On Wednesday, US heavyweight Rulon Gardner lost his Olympic title after losing in the semi-finals of the 120kg Greco-Roman wrestling event against Georgiy Tsurtsumia of Kazakhstan.}''\\
\hline
P2 & ``\textit{Following his loss to Georgiy Tsurtsumia of Kazakhstan in the semi-finals of the 120kg Greco-Roman wrestling event, US heavyweight Rulon Gardner lost his bid for repeat Wrestling Gold.}''\\
\hline
P3 & ``\textit{Rulon Gardner, the heavyweight world champion, suffered a major upset on Wednesday as he was defeated in the semi-finals of the 120kg Greco-Roman wrestling tournament by Georgiy Tsurtsumia of Kazakhstan. Gardner lost his Olympic title bid to {\color{red} George Kubica} {\color{red} after losing 1-1 and 6-1.}}''\\
\hline
P4 & ``\textit{US heavyweight Rulon Gardner suffered a setback in the {\color{red} final round} of his Olympic title against Georgiy Tsurtsumia of Kazakhstan, {\color{red} leading to an upset loss that could have resulted in injury.}}''\\
\hline
P5 & ``\textit{Despite winning the Olympic title {\color{red} in 2004}, Rulon Gardner, US heavyweight champion, lost his quest to repeat as a wrestler in the 120kg Greco-Roman wrestling event after losing to Georgiy Tsurtsumia of Kazakhstan in their semi-final match.}''\\
\hline
P6 & ``\textit{The US heavyweight champion, Rulon Gardner of the United States, suffered a defeat to Georgiy Tsurtsumia from Kazakhstan in the semi-finals of his men's 120kg Greco-Roman wrestling tournament, which cost him the Olympic championship on Wednesday.}''\\
\hline
P7 & ``\textit{In a bid to avoid repeat Wrestling Gold, US heavyweight Rulon Gardner lost his Olympic title on Wednesday after losing in the {\color{red} final round of Georges Tsurtsumia'S} 120kg Greco-Roman wrestling match against Kazakhstan's {\color{red} Rorum Ignite}.}''\\
\hline
P8 & ``\textit{{\color{red} After} losing his Olympic title on Wednesday, US heavyweight Rulon Gardner was defeated in the semi-finals of Greco–Roman wrestling event at 120kg against Georgiy Tsurtsumoj, Kazakhstan.}''\\
\hline
\end{tabular}
\label{example-one-paraphrase-without-riff}
\end{table*}

\begin{table*}
\small
\centering
\caption{Generated top eight paraphrases using diverse beam search \textbf{after fine-tuning the paraphrase model with the RIFF method} in the 128-shot setting. The example input sentence is from the AGNews test dataset:\\
``\textit{Gardner Loses Quest for Repeat Wrestling Gold US heavyweight Rulon Gardner lost his Olympic title Wednesday after being beaten in the semi-final stage of the 120kg Greco-Roman wrestling event by Georgiy Tsurtsumia of Kazakhstan.}'' Contradicting information or hallucinations are highlighted in {\color{red} red}}

\begin{tabular}{ p{0.1\linewidth} | p{0.87\linewidth} }
\hline
Paraphrase & Text \\
\hline
P1 & ``\textit{The US heavyweight Rulon Gardner lost his Olympic title on Wednesday after losing in the semi-final stage of the 120kg Greco-Roman wrestling event to Georgiy Tsurtsumia of Kazakhstan.}''\\
\hline
P2 & ``\textit{On Wednesday, Rulon Gardner lost his Olympic title after losing in the semi-final stage of the 120kg Greco-Roman wrestling event to Georgiy Tsurtsumia of Kazakhstan.}''\\
\hline
P3 & ``\textit{Georgiy Tsurtsumia of Kazakhstan defeated US heavyweight Rulon Gardner in the semi-final stage of the 120kg Greco-Roman wrestling event, resulting in Gardner losing his Olympic title on Wednesday.}''\\
\hline
P4 & ``\textit{{\color{red} Despite} losing his Olympic title on Wednesday, US heavyweight Rulon Gardner lost his quest for {\color{red} repeat wrestling}. He was defeated in the 120kg Greco-Roman wrestling event by Georgiy Tsurtsumia of Kazakhstan in their semi-final stage.}''\\
\hline
P5 & ``\textit{Rulon Gardner, the heavyweight US heavy weight loser of Olympic title after losing in the semi-final stage of the 120kg Greco-Roman wrestling event to Georgiy Tsurtsumia of Kazakhstan.}''\\
\hline
P6 & ``\textit{US heavyweight Rulon Gardner suffered a loss in the semi-final stage of the 120kg Greco-Roman wrestling event {\color{red} after} losing his Olympic title.}''\\
\hline
P7 & ``\textit{In the 120kg Greco-Roman wrestling event, Rulon Gardner lost his Olympic title on Wednesday after losing in the semi-final stage by Georgiy Tsurtsuma of Kazakhstan.}''\\
\hline
P8 & ``\textit{{\color{red} After} losing his Olympic title on Wednesday, Rulon Gardner lost the semi-final stage of a 120kg Greco–Roman wrestling event against Georgiy Tsurtsumoia of Kazakhstan in the Quest for {\color{red} Repeat Wrestling}.}''\\
\hline
\end{tabular}
\label{example-one-paraphrase-with-riff}
\end{table*}

\begin{table*}
\small
\centering
\caption{Generated top eight paraphrases using diverse beam search {\textbf{with the pre-trained paraphrase model (distilled from GPT3.5-turbo)}} for the following input sentence from the AGNews test dataset:\\
``\textit{Calif. Aims to Limit Farm-Related Smog (AP) AP - Southern California's smog-fighting agency went after emissions of the bovine variety Friday, adopting the nation's first rules to reduce air pollution from dairy cow manure.}'' Contradicting information or hallucinations are highlighted in {\color{red} red}.}

\begin{tabular}{ p{0.1\linewidth} | p{0.87\linewidth} }
\hline
Paraphrase & Text \\
\hline
P1 & ``\textit{The smog-fighting agency in southern California followed {\color{red} the lead of other states} by implementing the first rules to control air pollution from dairy cow manure emissions.}''\\
\hline
P2 & ``\textit{Southern California's smog-fighting agency took action on bovine eutrophication by adopting the first rules of the nation to reduce farm-related air pollution, including dairy cow manure emissions.}''\\
\hline
P3 & ``\textit{California's smog-fighting agency in southern California took action on bovine deposition Friday, following the implementation of the first rules in the country to control air pollution from dairy cow manure.}''\\
\hline
P4 & ``\textit{Calf Air Pollution Control: Southern California's smog-fighting agency took action on Friday to limit farm-related {\color{red} sewage} emissions by adopting the first rules in the country to reduce air pollution from dairy cow manure.}''\\
\hline
P5 & ``\textit{Efforts to Reduce Farm-Related Smog: Calif.'s agency in Southern California set the pace on Friday with its first-ever goals to combat bovine emissions and the {\color{red} nation'll be moving quickly} to eliminate farm-related air pollution, {\color{red} including dairy} cow manure.}''\\
\hline
P6 & ``\textit{Southern California's {\color{red} EPA} sought to reduce air pollution from farm-related manure, and the agency adopted the first rules {\color{red} in history} to address bovine smog emissions.}''\\
\hline
P7 & ``\textit{Calif. aims to reduce farm-related smog, which Southern California\'s {\color{red} first specialized} agency for reducing bovine emissions took action on Friday by adopting the {\color{red} nation' reviews} that address this issue with ""first ever"" rules from the {\color{red} U.S. federal government}.}''\\
\hline
P8 & ``\textit{The smog-fighting agency in Southern California adopted new regulations on Friday to reduce farm-related pollution, including air pollution from dairy cow manure.}''\\
\hline
\end{tabular}
\label{example-two-paraphrase-without-riff}
\end{table*}

\begin{table*}
\small
\centering
\caption{Generated top eight paraphrases using diverse beam search \textbf{after fine-tuning the paraphrase model with the RIFF method} in the 128-shot setting. The example input sentence is from the AGNews test dataset:\\
``\textit{Calif. Aims to Limit Farm-Related Smog (AP) AP - Southern California's smog-fighting agency went after emissions of the bovine variety Friday, adopting the nation's first rules to reduce air pollution from dairy cow manure.}'' Contradicting information or hallucinations are highlighted in {\color{red} red}}

\begin{tabular}{ p{0.1\linewidth} | p{0.87\linewidth} }
\hline
Paraphrase & Text \\
\hline
P1 & ``\textit{The smog-fighting agency in southern California adopted the first rules to reduce air pollution from dairy cow manure, targeting bovine emissions.}''\\
\hline
P2 & ``\textit{Southern California's smog-fighting agency took action on bovine emissions by adopting the nation' first rules to reduce air pollution from dairy cow manure.}''\\
\hline
P3 & ``\textit{On Friday, Southern California's smog-fighting agency adopted the nation' first rules to reduce air pollution from dairy cow manure emissions.}''\\
\hline
P4 & ``\textit{California's smog-fighting agency took action on bovine emissions by adopting the nation' first rules to reduce air pollution from dairy cow manure.}''\\
\hline
P5 & ``\textit{Calif. seeks to reduce farm-related smog emissions by adopting the nation's first rules on Friday, which targets bovine emission in Southern California and is {\color{red} one of the country' best agencies} for curbing air pollution from dairy cow manure.}''\\
\hline
P6 & ``\textit{Southern California's scounding agency took action on bovine emissions by adopting the nation' first rules to reduce air pollution from dairy cow manure.}''\\
\hline
P7 & ``\textit{The smog-fighting agency of Southern California targeted bovine emission on Friday, adoptting the country's first rules to reduce air pollution from dairy cow manure.}''\\
\hline
P8 & ``\textit{AP: Southern California's {\color{red} federal} agency adopting the nation' first rules to reduce air pollution from dairy cow manure.}''\\
\hline
\end{tabular}
\label{example-two-paraphrase-with-riff}
\end{table*}

\section{Further Training Details}
\label{training-details-extra:appendix}

The learning rate for each LM tuning technique was separately fine-tuned from the set \{0.5, 0.3, 0.1, 0.01, 0.001, 0.0001, 0.00001\} using the train/validation split created for the seed 11 on the SST2 dataset. The tuned learning rates were then applied globally across other datasets and experiments. For paraphrase fine-tuning, we train all the parameters in T5-base with the learning rate of 0.00001. In Tables~\ref{hyper-parameters} and~\ref{learning-rates}, we list the hyper-parameters and learning rates used across all datasets.  For optimization, we utilized the AdamW~\cite{DBLP:journals/corr/abs-1711-05101}\footnote{\tiny \url{https://pytorch.org/docs/stable/generated/torch.optim.AdamW.html}} optimizer with the AMSGrad variant set to True~\cite{DBLP:journals/corr/abs-1904-09237}. We implemented the methods using the HuggingFace\footnote{\tiny \url{https://huggingface.co/}} library and the PyTorch\footnote{\tiny \url{https://pytorch.org/}} machine learning framework. We report the accuracy metric on these classification datasets. The experiments were conducted using multiple NVIDIA's A40 GPU cards.

\begin{table}[t]
\centering
\caption{Shared hyper-parameters used across all experiments and datasets.}
\begin{tabular}{ c | c }
\hline
Hyper-parameter & Value\\
\hline
Top-$k$ candidates in \textit{\textit{\textit{GS}}} & $k$=4 \\
batch size (RoBERTa-large) & 8 \\
batch size in \textit{\textit{\textit{GS}}} (RoBERTa-large) & 2 \\
Weight decay & 0.0001 \\
Max epochs & 100 \\
length cutoff & 128 tokens \\
Paraphrase sample size & $M$=8 \\
Checkpointing steps & 8 \\
$D^{'}$ in \textit{ClsTune} & 128 \\
Prompt len in \textit{SpTune} & $L$=25 \\
$\beta$ in MML & 0.1 \\
$\beta$ in PG & 0.6 \\
\textit{LoRA} $\alpha$ & 32 \\
\textit{LoRA} $r$ & 8 \\
\textit{LoRA} dropout & 0.1 \\
Diversity penalty for Div beam & 3.0 \\
Repetition penalty for Div beam & 10.0 \\
Temperature in Div beam & 0.7 \\
P value for top-p & 0.99
\end{tabular}
\label{hyper-parameters}
\end{table}

\begin{table}[t]
\centering
\caption{Learning rates used per Language Model (LM) tuning technique.}
\begin{tabular}{ c | c }
\hline
LM Tuning Technique & Learning Rate\\
\hline
\textit{\textit{\textit{\textit{GS}}}} & No rate \\
\textit{\textit{AllTune}} & 0.00001 \\
\textit{\textit{InTune}} &  0.001 \\
\textit{\textit{HTune}} & 0.001 \\
\textit{\textit{ClsTune}} & 0.001 \\
\textit{\textit{SpTune}} & 0.001 \\
\textit{\textit{LoRA}} & 0.0001
\end{tabular}
\label{learning-rates}
\end{table}

\section{Extended Related Works}
\label{related-works-extra:appendix}

\noindent
{\bf Prompt Optimization \& Efficient Tuning}: Recent research proposes various techniques for prompt optimization and efficient tuning of language models. In our experiments, we have used successful techniques from each of these areas.

FluentPrompt~\cite{shi2022human} is a recent discrete prompting technique based on the projected gradient-descent and Langevin dynamics. FluentPrompt introduces a fluency constraint within Langevin dynamics to generate a sample of high-performing prompts for more interpretable analysis of these discrete prompts. The optimized prompts by FluentPrompt performs on-par to the AutoPrompt, however they have lower perplexity~\cite{shi2022human}.

Building upon \textit{SpTune}~\cite{lester-etal-2021-power} and P-tuning~\cite{li-liang-2021-prefix}, P-tuning V2~\cite{liu-etal-2022-p} introduced the concept of deep prompt tuning. This method involves injecting prompt vectors into the deeper layers of the transformer model to close the performance gap with AllTuning in medium-sized language models. We have experimented with \textit{LoRA}~\cite{DBLP:journals/corr/abs-2106-09685}, a recent low-rank adaptation technique for tuning language models. Other potential methods include training bottleneck adapter modules~\cite{DBLP:journals/corr/abs-1902-00751, lin-etal-2020-exploring} added per sub-layer of the transformer model. \textit{LoRA} outperforms adapter tuning and P-Tuning V2 techniques ~\cite{DBLP:journals/corr/abs-2106-09685}. The successors of \textit{LoRA} include DyLoRA~\cite{valipour-etal-2023-dyLoRA}  which dynamically learns a range of adaptation ranks, thus eliminating the need to search the rank of the adaptation matrices as a hyper-parameter. Similarly, Ada\textit{LoRA} dynamically allocates the parameter budget among the weight matrices during adaptation, with matrices of higher priority (i.e., those with greater importance to the downstream task) receiving higher adaptation ranks than less important matrices~\cite{zhang2023adaptive}.

In scenarios where gradients are absent, Black-Box Tuning~\cite{DBLP:journals/corr/abs-2201-03514} applies derivative-free algorithms for optimizing continuous prompts. For discrete prompt optimization, RLPrompt~\cite{deng-etal-2022-rlprompt} employs the on-policy version of soft Q-learning~\cite{https://doi.org/10.48550/arxiv.2106.07704} to find the optimal prompt tokens in a gradient-free setting. Decoder Tuning~\cite{cui-etal-2023-decoder} learns a decoder network over the language model, thus circumventing the need for gradient computation and input-side prompt tuning in few-shot classification. In a recent study, TEMPERA~\cite{zhang2022tempera} introduced a novel approach that involves test-time discrete prompt editing using a trained RL agent. This agent is capable of modifying the instruction, in-context examples, or the verbalizers based on the given task input.

The use of Language Models (LLMs) in generating instructions for downstream tasks has involved a two-step process. Initially, LLMs generate a set of candidate instructions, and subsequently, the highest-scoring instruction is utilized to prompt another LLM to perform the downstream task. This approach, known as prompt-based generation-then-filtering, has been investigated in the recent APE method~\cite{zhou2023large}. APE demonstrates the ability to generate prompts that achieve performance comparable to human-designed prompts~\cite{zhou2023large}.

To prompt language models for reasoning tasks, another line of research augment the input context with demonstration examples outlining the intermediate reasoning steps to form the answer. Providing manually or automatically generated chain-of-thoughts within these demonstrations strikingly improve LLMs performance in reasoning tasks~\cite{DBLP:journals/corr/abs-2201-11903, zhang2022automatic, NEURIPS2022_8bb0d291}.

All of the aforementioned techniques for prompt optimization and efficient tuning of the language model use the original task's input text (or the original input context) provided within the dataset.

\noindent
{\bf RL for Paraphrase Generation}:
In the following paragraphs, we provide a brief overview of similar reinforcement learning objectives employed for paraphrase generation. Li et al.~\cite{li-etal-2018-paraphrase} used a deep RL technique, training a pointer-generator network as the paraphrase generator and a decomposable attention model as the evaluator which assigns a paraphrase score to pairs of sentences. The generator was trained using the policy gradient objective, with reward shaping and scaling to stabilize the training process~\cite{li-etal-2018-paraphrase}. Another approach by Qian et al.~\cite{qian-etal-2019-exploring} focused on generating diverse paraphrases by training multiple generators, accompanied by a paraphrase discriminator and a generator discriminator. Policy gradient objective and self-critical learning~\cite{DBLP:journals/corr/RennieMMRG16} were employed for training the generators, with the baseline reward used in the policy gradient objective being the reward obtained from the greedy-decoded sequence. Liu et al.~\cite{liu-etal-2020-learning} also applied the policy gradient objective with self-critical learning, incorporating multiple reward functions such as Rouge score with the reference paraphrase, negative Rouge score with the input sentence to encourage lexical variations, and semantic similarity score between the paraphrase and the input sentence to ensure semantic fidelity.

Another study by Du and Ji~\cite{du-ji-2019-empirical} compared the use of imitation learning algorithm DAGGER with policy gradient REINFORCE for paraphrase generation. The policy gradient objective has also been applied in generating paraphrases while considering multiple objectives for entailment relation-aware paraphrase generation~\cite{Sancheti_Srinivasan_Rudinger_2022}. In the context of chatbot responses, a recent work studies unsupervised paraphrase generation with proximal policy optimization, aiming to maximize a combination of rewards such as textual entailment, semantic similarity, language fluency, and lexical dissimilarity~\cite{DBLP:journals/corr/abs-2103-12777}. Similarly, the policy gradient objective has been employed to optimize multiple rewards, similar to previous work, for unsupervised paraphrase generation~\cite{10.1145/3394486.3403231}.

While previous studies have applied RL techniques for paraphrase generation, we propose the use of MML gradients instead of policy gradients to train our paraphrase model. Our training objective fine-tunes the paraphrase model for a downstream classification task. Our paraphrase model has been distilled from a large language model.

\section{Task Instructions \& Input Format}
\label{task-instruct-input-format:appendix}
Table~\ref{task-input-format} provides a summary of the task instructions that we append before the inputs, as well as the class verbalizers for classifying the input text. The instructions and input templates are derived from prior work in prompt optimization~\cite{deng-etal-2022-rlprompt}.

\begin{table*}
\centering
\caption{Number of classes $C$, test set size $T$, the input format, and the instruction used per dataset. The label words are provided within the instructions.}
\begin{tabular}{ p{0.1\linewidth} | p{0.02\linewidth} | p{0.05\linewidth} | p{0.2\linewidth} | p{0.5\linewidth} }
\hline
Dataset & $C$ & $T$ & Input Format & Instruction \\
\hline
SST2 & 2 & 1821 & \small ``<s> \{Instruction\} \{Text\} . It was <mask> . </s>'' & \small ``In this task, you are given sentences from movie reviews. The task is to classify a sentence as `great' if
the sentiment of the sentence is positive or as `terrible' if the sentiment of the sentence is negative.''\\
\hline
SST5 & 5 & 2210 & \small ``<s> \{Instruction\} \{Text\} . It was <mask> . </s>'' & \small ``In this task, you are given sentences from movie reviews. Based on the given review, classify it to one of
the five classes: (1) terrible, (2) bad, (3) okay, (4) good, and (5) great.''\\
\hline
CR & 2 & 2000 & \small ``<s> \{Instruction\} \{Text\} . It was <mask> . </s>'' & \small ``In this task, you are given sentences from customer reviews. The task is to classify a sentence as `great' if
the sentiment of the sentence is positive or as `terrible' if the sentiment of the sentence is negative.''\\
\hline
MR & 2 & 2000 & \small ``<s> \{Instruction\} \{Text\} . It was <mask> . </s>'' & \small ``In this task, you are given sentences from movie reviews. The task is to classify a sentence as `great' if
the sentiment of the sentence is positive or as `terrible' if the sentiment of the sentence is negative.''\\
\hline
TREC & 6 & 500 & \small ``<s> \{Instruction\} <mask>: \{Text\} . </s>'' & \small ``You are given a question. You need to detect which category better describes the question. Answer with `Description', `Entity', `Expression', `Human', `Location', and `Number'.'' \\
\hline
AG's News & 4 & 7600 & \small ``<s> \{Instruction\} <mask> News: \{Text\} . </s>'' & \small ``In this task, you are given a news article. Your task is to classify the article to one out of the four topics
`World', `Sports', `Business', `Tech' if the article's main topic is relevant to the world, sports, business,
and technology, correspondingly. If you are not sure about the topic, choose the closest option.''\\
\hline
\end{tabular}
\label{task-input-format}
\end{table*}

\end{document}